\documentclass[10pt,twocolumn,letterpaper]{article}
\usepackage{graphicx}
\usepackage[datasets]{wacv}

\newcommand{\name}{ABACUS}

\usepackage{booktabs}
\usepackage{multirow}
\usepackage{amsmath,amssymb}
\usepackage{xcolor}
\usepackage{pifont}
\usepackage{afterpage}

\newcommand{\cmark}{\textcolor{green!55!black}{\ding{51}}}
\newcommand{\xmark}{\textcolor{red!70!black}{\ding{55}}}
\newcommand{\pmark}{\textcolor{orange!85!black}{$\sim$}}

\definecolor{wacvblue}{rgb}{0.21,0.49,0.74}
\usepackage[pagebackref,breaklinks,colorlinks,allcolors=wacvblue]{hyperref}

\def\wacvPaperID{2002}
\def\confName{WACV}
\def\confYear{2027}

\title{\name{}: A Benchmark for Image Editing Models to Assess Object
Category and Counting Under Scene Conservation}

\author{Aymen Lassoued \quad Angelique Loesch \quad Romaric Audigier\\
Universit\'e Paris-Saclay, CEA, List, F-91120 Palaiseau, France\\
{\tt\small \{firstname.lastname\}@cea.fr}
}

\begin{document}
\maketitle

\begin{abstract}
Instruction-based image editing models can now add, remove a specified
number of objects or change their category while preserving the scene.
Yet these models often fail to produce the correct number, and these
failures grow sharply as scenes become more complex and the requested
count increases. Current benchmarks largely miss this problem: they score
edits with holistic similarity metrics or with vision-language model
(VLM) judges that are themselves poor at counting, so numerical errors go
unnoticed or are graded incorrectly. As a result, the true numerical
ability of today's editing models has never been accurately measured.

We address this issue with \name{}, a new benchmark, evaluation protocol and
dedicated scoring metric for numerical image editing. Our protocol is
\emph{judge-free}: no VLM or MLLM scores an edit or counts objects (a VLM only confirms the category of rare ambiguous crops).
Instead, it verifies each edit at the instance level against panoptic ground truth
confirmed by detector consensus, checking both that the exact requested
count was achieved and that the rest of the scene was preserved, across
the add, remove and change operations. By evaluating leading image
editors, our method reveals substantial numerical editing failures that
worsen with increasing scene complexity and object count; failures that
prior benchmarks often overlooked or misjudged.
\end{abstract}

\section{Introduction}
\label{sec:intro}

\begin{figure}[t]
\centering
\includegraphics[width=0.92\linewidth]{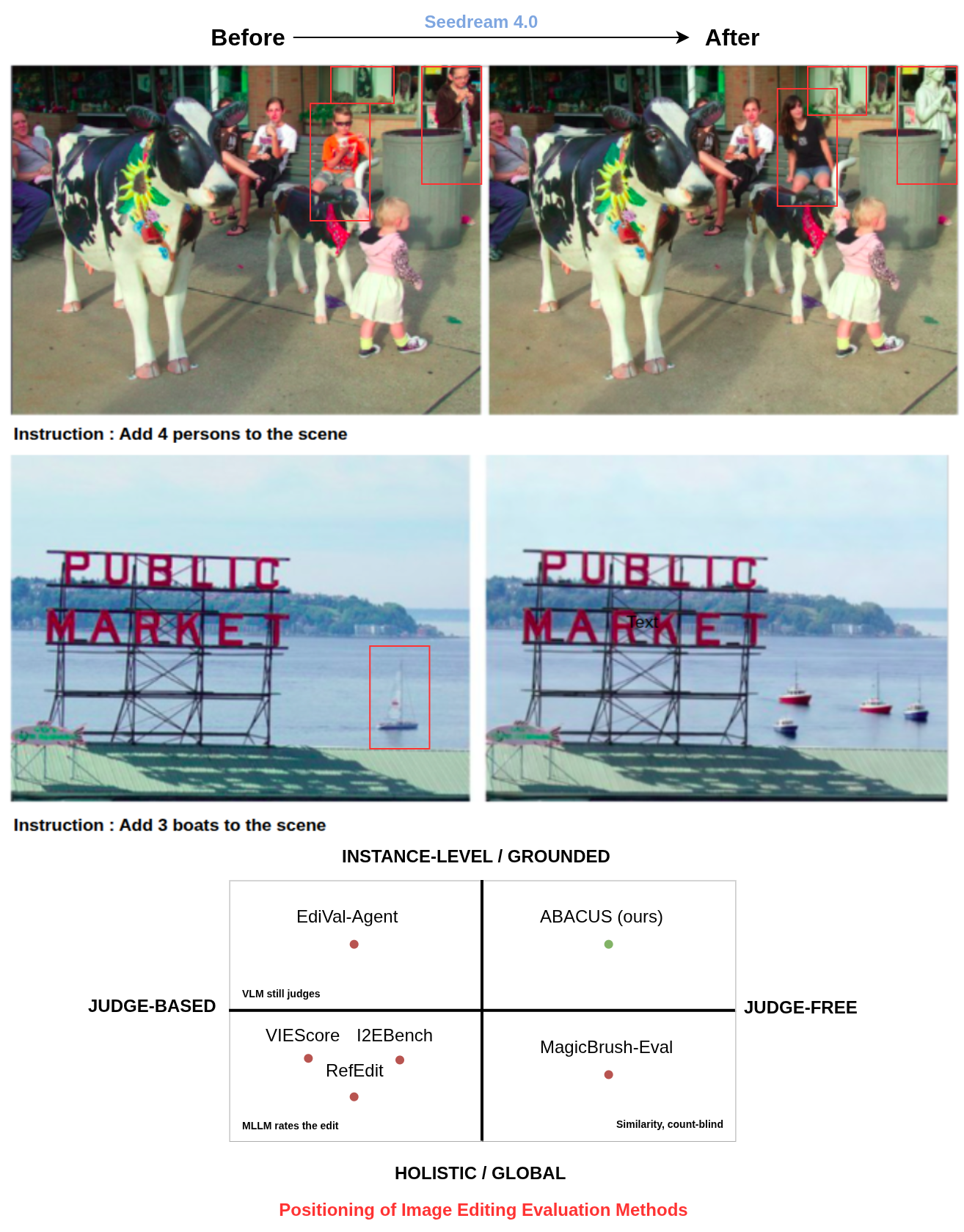}
\caption{
\textbf{Numerical editing requires a specialized evaluator.}
Given an instruction to generate an exact number of objects, existing evaluation methods, including VLM judges (VIEScore~\cite{viescore}, I2EBench~\cite{i2ebench}, and RefEdit~\cite{refedit}) and the agentic evaluator EdiVal-Agent~\cite{edival}, incorrectly mark the edit as successful because they fail to verify numerical correctness. In contrast, \name{} correctly identifies the edit as a failure. (The red boxes indicate the target categories that were either modified or removed during editing.)
}
\label{fig:teaser}
\end{figure}

Instruction-based image editing has advanced rapidly: a single model can
now add, remove, or change objects in a real image from a short
natural-language command. Proprietary systems such as Gemini~2.5 Flash
Image (``Nano~Banana'')~\cite{geminiflash}, Seedream~4.0~\cite{seedream4},
and GPT-Image-1~\cite{gptimage1}, together with strong open models such as
Qwen-Image-Edit~\cite{qwenimage}, FLUX.1~Kontext~\cite{fluxkontext},
Step1X-Edit~\cite{step1xedit}, and BAGEL~\cite{bagel}, produce edits that
are visually convincing and, at a glance, faithful to the instruction.
However, producing the \emph{exact} number of objects an instruction
requests (\emph{numerical} editing), as in ``add three boats'', ``remove
two cars'', or ``change four of the cows into sheep'', is quite hard.
The model must produce the right count, and it must do so
\emph{without disturbing the rest of the scene}: it should not silently add
an unrequested object, delete an existing one, or repurpose an original
instance to fake the count. We find that editors frequently fail at this
(Fig.~\ref{fig:teaser}), and that the failures grow sharply as scenes become more crowded and
the requested count increases, precisely the regime where reliable counting
matters most.

Measuring this matters beyond the editor. Generative editing is increasingly
used to build \emph{controlled training data}~\cite{augsurvey,soc}: when real
images of a configuration are scarce (rare counts, dense crowds, long-tail
categories), an editor renders a set number of instances, and one silent
miscount corrupts every downstream label. Being judge-free, deterministic, and
grounded in detected evidence, the same signal could also serve as a \emph{verifiable
reward} (RLVR) for post-training editors toward exact counting, where a learned
VLM reward would be noisy and reward-hackable.

The deeper problem is that \emph{we cannot currently measure this failure}:
the instruments used to evaluate editing are themselves unreliable for
counting (Sec.~\ref{sec:related}). Holistic similarity metrics
(CLIP~\cite{clip}, DINO~\cite{dino}, $L_1/L_2$, FID~\cite{fid}), as in
MagicBrush~\cite{magicbrush}, capture global appearance but are blind to
instance counts. The now-dominant alternative scores edits with a VLM or MLLM
\emph{judge}~\cite{viescore,i2ebench,imgedit,krisbench,giebench,compbench},
which is circular: VLMs are themselves poor counters. Their accuracy collapses
with object number and variety~\cite{vlmcount20,vlmcountsynthetic,makeitcount,t2icannotcount},
they miss small edits~\cite{viescore,mllmjudge}, score
unreliably~\cite{mllmperceptualbias,vlmjudgescore}, and have no notion of
\emph{instance correspondence}: when a model turns one of two people into a
third, or deletes the sources and paints the targets elsewhere, the judge
still reports success (Fig.~\ref{fig:teaser}). Editors' numerical ability
has thus been overlooked or misjudged.

We argue that numerical editing must be \emph{verified}, not judged. We
introduce \name{}, a benchmark, an evaluation protocol, and a dedicated
scoring metric for numerical image editing. \name{} is built on COCO
images~\cite{coco} with high-quality panoptic ground truth from
COCONut~\cite{coconut}, and it scores every edit by reconstructing
\emph{what happened to each object instance} rather than by asking a model
for an opinion. It segments every instance in both images~\cite{sam3},
matches before/after instances with a matcher robust to generative appearance
drift, and confirms each category by the \emph{independent agreement} of two
detectors~\cite{rfdetr,codetr}. From this, it assigns every object a verdict
(\emph{kept}, \emph{added}, \emph{removed}, \emph{transformed}) and jointly
checks the \emph{exact requested count} and \emph{scene conservation} across
add, remove, and change. No VLM or MLLM ever judges an edit or counts objects.

\name{} covers \emph{add}, \emph{remove}, and \emph{change} edits over
COCO/COCONut~\cite{coco,coconut} with instance-level ground truth, exact target
counts, and easy/medium/hard difficulty tiers (Sec.~\ref{sec:difficulty}). Its counts are confirmed
by detector consensus and \emph{every} sample is human-verified, both that the
requested edit is feasible and that its difficulty tier is correct, so the
benchmark is reliable by construction. We release it
publicly, with a datasheet and license (App.~\ref{sec:appendix-datasheet}).\footnote{Code and data: \href{https://github.com/CEA-LIST/ABACUS}{github.com/CEA-LIST/ABACUS}}

Validated against human annotation, \name{}'s automatic verdicts closely
track human counting across all difficulty tiers and
seven editors that span a wide range of edit quality,
\emph{including the hard, high-count regime where VLM judges collapse}.
Using \name{} to evaluate leading editors, we expose substantial and
systematic numerical failures that prior benchmarks had hidden or misjudged.
These failures intensify with scene complexity and target count.

\section{Related Work}
\label{sec:related}

\subsection{Instruction-based image editing}
Editing from natural-language instructions was popularized by
InstructPix2Pix~\cite{instructpix2pix} and the manually annotated
MagicBrush~\cite{magicbrush} dataset. Quality has since risen sharply with
unified and in-context models. On the proprietary side, Gemini~2.5 Flash
Image~\cite{geminiflash}, Seedream~4.0~\cite{seedream4}, and
GPT-Image-1~\cite{gptimage1} set the current bar for targeted,
language-driven edits; on the open side, Qwen-Image-Edit~\cite{qwenimage},
FLUX.1~Kontext~\cite{fluxkontext}, Step1X-Edit~\cite{step1xedit}, and
BAGEL~\cite{bagel} approach proprietary quality. All of these can
\emph{attempt} numerical edits; whether they achieve the exact requested
count while preserving the scene is what \name{} measures, and what
their own benchmarks do not.
\begin{table*}[t]
  \centering
  \caption{\textbf{Overview of \name{}.}}
  \label{tab:overview}

  \begin{subtable}[t]{0.64\textwidth}
    \centering
    \resizebox{\textwidth}{!}{\setlength{\tabcolsep}{3pt}
    \renewcommand{\arraystretch}{0.9}
    \begin{tabular}{l cc c cccc}
      \toprule
      & \multicolumn{2}{c}{Data \& supervision} & Task & \multicolumn{4}{c}{Evaluation protocol} \\
      \cmidrule(lr){2-3}\cmidrule(lr){4-4}\cmidrule(lr){5-8}
      Benchmark & Inst.\ GT & GT masks & Multi-inst. & Judge-free & Mask pres. & Num.\ metric & Grounded count \\
      \midrule
      EditVal~\cite{editval2023}      & \cmark & \xmark & \xmark & \xmark & \xmark & \xmark & \xmark \\
      MagicBrush~\cite{magicbrush}    & \xmark & \pmark & \pmark & \cmark & \xmark & \xmark & \xmark \\
      ImgEdit-Bench~\cite{imgedit}    & \xmark & \xmark & \pmark & \xmark & \xmark & \xmark & \xmark \\
      I2EBench~\cite{i2ebench}        & \xmark & \xmark & \cmark & \xmark & \xmark & \cmark & \xmark \\
      CompBench~\cite{compbench}      & \xmark & \xmark & \pmark & \xmark & \xmark & \xmark & \xmark \\
      KRIS-Bench~\cite{krisbench}     & \xmark & \xmark & \cmark & \xmark & \xmark & \cmark & \xmark \\
      RefEdit-Bench~\cite{refedit}    & \cmark & \pmark & \xmark & \xmark & \xmark & \xmark & \xmark \\
      ICE-Bench~\cite{icebench}       & \xmark & \xmark & \pmark & \xmark & \xmark & \xmark & \xmark \\
      VIBE~\cite{vibe}                & \xmark & \xmark & \xmark & \xmark & \xmark & \xmark & \xmark \\
      GIE-Bench~\cite{giebench}       & \pmark & \cmark & \xmark & \xmark & \cmark & \xmark & \xmark \\
      EdiVal-Agent~\cite{edival}      & \xmark & \xmark & \cmark & \xmark & \xmark & \cmark & \xmark \\
      \textbf{\name{} (ours)}         & \cmark & \cmark & \cmark & \cmark & \cmark & \cmark & \cmark \\
      \bottomrule
    \end{tabular}}
    \caption{\textbf{Positioning vs.\ prior benchmarks.} \cmark{} full,
    \pmark{} partial, \xmark{} absent. \emph{Inst.\ GT}: instance segmentation;
    \emph{GT masks}: masks separating edited vs.\ preserved regions;
    \emph{Multi-inst.}: prompts specify $K\!>\!1$ instances; \emph{Judge-free}:
    no VLM/MLLM scores the edit (a VLM may assist in a non-scoring role, as in
    ours); \emph{Mask pres.}: mask-based preservation on non-edited regions;
    \emph{Num.\ metric}: an explicit object-count metric; \emph{Grounded count}:
    the object count comes from detection evidence cross-checked by detector
    agreement. \name{} is the only benchmark that is judge-free
    \emph{and} grounded on numerality.}
    \label{tab:benchmarks}
  \end{subtable}\hfill
  \begin{subtable}[t]{0.34\textwidth}
    \centering
    \resizebox{\textwidth}{!}{\setlength{\tabcolsep}{4pt}
    \begin{tabular}{lcccc}
      \toprule
       & Add & Remove & Change & Total \\
      \midrule
      Easy   & 1{,}159 & 1{,}146 & 1{,}000 & 3{,}305 \\
      Medium & 1{,}363 & 1{,}340 & 1{,}128 & 3{,}831 \\
      Hard   & 1{,}374 & 1{,}308 & 1{,}126 & 3{,}808 \\
      \midrule
      Total    & 3{,}896 & 3{,}794 & 3{,}254 & \textbf{10{,}944} \\
      mean $K$ & 3.20 & 2.32 & 2.30 & \\
      max $K$  & 15 & 10 & 10 & \\
      \bottomrule
    \end{tabular}}
    \caption{\textbf{\name{} composition.} Number of instructions per operation and
    difficulty tier, over 3{,}845 COCO source images (2.85 instructions each)
    spanning all 80 categories ($K\!=\!1$--$15$, 72 change pairs).
    mean $K$ and max $K$: mean and maximum requested count.}
    \label{tab:composition}
  \end{subtable}
\end{table*}
\subsection{Benchmarks and metrics for image editing}
Early evaluation relied on holistic similarity metrics (CLIP~\cite{clip}, DINO~\cite{dino}, $L_1$,
FID~\cite{fid}), as used to score MagicBrush~\cite{magicbrush} edits, which are
insensitive to object count. The field has since moved to VLM/MLLM
\emph{judges}. VIEScore~\cite{viescore} prompts an MLLM for an explainable
score but correlates only weakly with humans and is itself noted to miss
small edits. I2EBench~\cite{i2ebench}, ImgEdit-Bench~\cite{imgedit},
KRIS-Bench~\cite{krisbench}, and CompBench~\cite{compbench} score edits
across many dimensions with MLLM judges. Notably, I2EBench and KRIS-Bench
even include a dedicated counting dimension, yet they grade it with a VLM
(GPT-4V / GPT-4o), inheriting exactly the counting unreliability we
document. GEdit-Bench~\cite{step1xedit} grades against real-world
instructions. RefEdit~\cite{refedit} contributes
RefEdit-Bench (rooted in RefCOCO) for referring-expression edits and reports
that editors handling a single prominent object degrade sharply once a scene
contains multiple entities, reinforcing our focus on scene complexity. And
learned scorers such as ADIEE~\cite{adiee} and human-aligned MLLM
judges~\cite{mllmjudge} try to repair judge reliability. All keep a model in the scoring loop for a task
that model performs poorly, and none target \emph{exact count} with
instance correspondence. Two efforts move toward grounding.
GIE-Bench~\cite{giebench} adds object-aware conservation masks but still
verifies functional correctness with VLM-answered multiple-choice
questions. EdiVal-Agent~\cite{edival} is the closest in spirit: it is
object-centric and reports that detectors combined with a VLM agree with
humans better than a VLM alone, which corroborates our premise. It differs
from \name{} in both goal and mechanism. EdiVal-Agent measures multi-turn
\emph{instruction following}: for symbolic edit types, including a
count-change task, it scores with a single open-vocabulary detector, but for
semantic types it routes detector-localized crops to a VLM and it generates
its instructions and object pool with a VLM, so the framework is judge-based
overall. Notably, its counts are read from one detector on arbitrary
images with no verified ground-truth reference. \name{} instead targets
\emph{exact-count fidelity and scene conservation}, grounding every verdict in
panoptic ground truth, segmentation, geometric matching, and two-detector
agreement, and never asking a VLM to count objects or judge whether an edit succeeded. Uniquely, it repairs
incomplete ground truth and checks that a ``change'' transformed the
\emph{same} instances \emph{in place}. Table~\ref{tab:benchmarks}
summarizes this positioning: no prior benchmark provides a fully
judge-free, instance-level protocol that jointly verifies exact count and
conservation, which is the gap \name{} fills.

Concretely, \name{} grounds counting in detection and segmentation. We build on
COCO~\cite{coco} with COCONut panoptic masks~\cite{coconut}, which are strong
enough to anchor instance-level verification while our pipeline is built to
tolerate their residual faults, and on strong detectors~\cite{rfdetr,codetr}
whose occasional errors are absorbed downstream by detector agreement,
presence-gated segmentation~\cite{sam3}, and the matching and
ground-truth-completion steps. The resulting verdicts are verifiable rather
than opined.
\begin{figure*}[t]
  \centering
  \includegraphics[width=\linewidth]{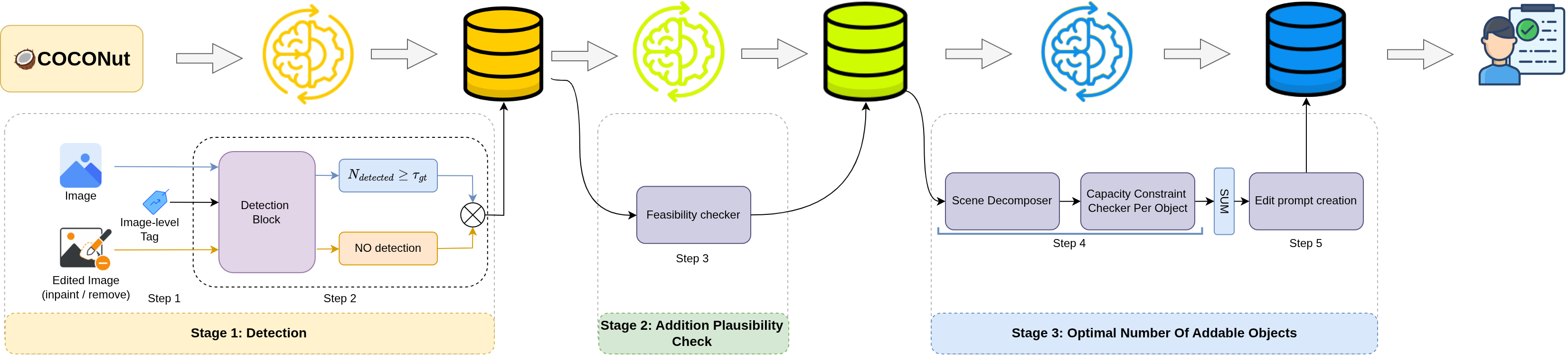}
  \caption{\textbf{Benchmark creation pipeline.} Each
  \{image, category\} pair from COCONut panoptic ground truth: a
  detector-verified removability audit, an addition-plausibility check, an
  optimal addable-count estimate, instruction generation, and a final
  human-verification pass
  (Sec.~\ref{sec:construction}).}
  \label{fig:pipeline}
\end{figure*}

\section{The \name{} benchmark}
\label{sec:benchmark}

\name{} is a benchmark: every edit is paired with a known target count and
a reliable instance-level reference. These are produced by a multi-stage
\emph{creation} pipeline that ends in human verification, and which is
entirely separate from the \emph{evaluation} protocol of
Sec.~\ref{sec:method}.

\subsection{Creation pipeline}
\label{sec:construction}
The pipeline (Fig.~\ref{fig:pipeline}) proceeds in five steps.

\noindent\textbf{1. Source and ground-truth counts.} We start from COCO
images~\cite{coco} with COCONut panoptic annotations~\cite{coconut}. For
each \{image, thing-category\} pair, the panoptic segments give the
ground-truth instance count $N_{\mathrm{GT}}$.

\noindent\textbf{2. Detector-verified removability audit.} For every pair
we run both detectors, RF-DETR~\cite{rfdetr} and Co-DETR~\cite{codetr}, on
the original image and keep it only if \emph{both} detect at least
$N_{\mathrm{GT}}$ instances of the category, so the ground-truth count is
independently confirmed rather than merely assumed. We then inpaint the
category region out of the image with a classical, non-generative method
(Navier-Stokes inpainting~\cite{navierstokesinpaint}) and run both detectors again; the pair passes
only if \emph{both} now detect zero instances. A pair that survives this
audit therefore has a detector-confirmed count and a region that is real
and can be cleanly removed, which makes its edits well posed.

\noindent\textbf{3. Addition plausibility check.} On the surviving pairs, a
vision-language model (Qwen3.6-27B~\cite{qwen36vl}) checks whether the scene plausibly admits the edit
(for an addition, whether there is a visible location where one more
instance of the category could be inserted), and pairs judged implausible
are dropped.

\noindent\textbf{4. Optimal number of addable objects.} For the plausible
pairs, the same model estimates the largest number of additional instances
that would look natural, decomposing the scene and checking a per-object
capacity constraint given the element scale and context,
yielding a per-image upper bound $K_{\max}$ on the target count.

\noindent\textbf{5. Instruction generation.} For removal and change we draw the target count $K$ as a uniform random integer in $\{1,\dots,N_{\mathrm{GT}}\}$, where $N_{\mathrm{GT}}$ is the known ground-truth count of the target category; for addition, where the addable count is not given a priori, we take the optimal addable-count estimate $K=K_{\max}$ of Step~4. We render a natural-language instruction for the operation, for example ``add $K$ \{category\} to the scene.'', ``change $K$ \{source\} into \{target\} in the scene.'', or ``delete $K$ \{category\} from the scene.'' with correct pluralization.
The instructions are intentionally simple, probing the counting ability alone
rather than spatial reasoning or attribute binding.

\noindent\textbf{Human verification.} Finally, \emph{every} sample is
inspected by two experienced annotators, with a final verifier resolving
disagreements, who confirm that the requested edit is
feasible on its image and that its difficulty tier is correct, and discard
any sample that fails either check.

The ground-truth counts come from panoptic annotation confirmed by the
dual-detector audit, and the vision-language model is used \emph{only} at
creation time to propose plausible instructions and count ranges; it never
sets a ground-truth count and never scores an edit. The dual-detector audit
likewise runs only on the \emph{source} images at creation time, disjoint
from the evaluation of any edited image (Sec.~\ref{sec:method}), so benchmark
selection cannot inflate a model's score; scoring is robust to which detectors
are used (App.~\ref{sec:appendix-robustness}). Combined with the
human-verification pass, this makes \name{} reliable by construction.

\subsection{Operations, difficulty, and scene complexity}
\label{sec:difficulty}

\name{} covers three operations: adding $K$ objects of a category, removing
$K$ of them, and changing $K$ instances of one category into another (one
example of each in Fig.~\ref{fig:examples} of App.~\ref{sec:appendix}). Rather than bin
difficulty by a single attribute, we grade it
with a composite
score built from three complementary axes, each rank-normalized to $[0,1]$
over the pool: the \emph{foreground coverage} of the scene (fraction of
pixels that are foreground), the \emph{target count} (the maximum number of
target instances the scene admits, which bounds the requested $K$), and the
\emph{target-object scale} (the share of the foreground occupied by the
target category, target-class pixels over foreground pixels, where a smaller
share is harder). Their sum gives a difficulty score in $[0,3]$ that we
split at its terciles into easy, medium, and hard tiers
(examples in Fig.~\ref{fig:difficulty} of App.~\ref{sec:appendix}). The split is meaningful rather than arbitrary,
and non-circular: scene attributes that do \emph{not} enter the score (the
number of instances, the number of categories, and the crowd ratio) still
separate the tiers, and model performance decreases monotonically across them
(Sec.~\ref{sec:leaderboard} and Table~\ref{tab:leaderboard}; validated in App.~\ref{sec:appendix}). Overall, \name{} comprises $10{,}944$ instructions over $3{,}845$ COCO images, roughly balanced across the three tiers and the add, remove, and change operations (Table~\ref{tab:composition}).

\section{Judge-free evaluation protocol}
\label{sec:method}

\begin{figure*}[t]
  \centering
  \includegraphics[width=\linewidth]{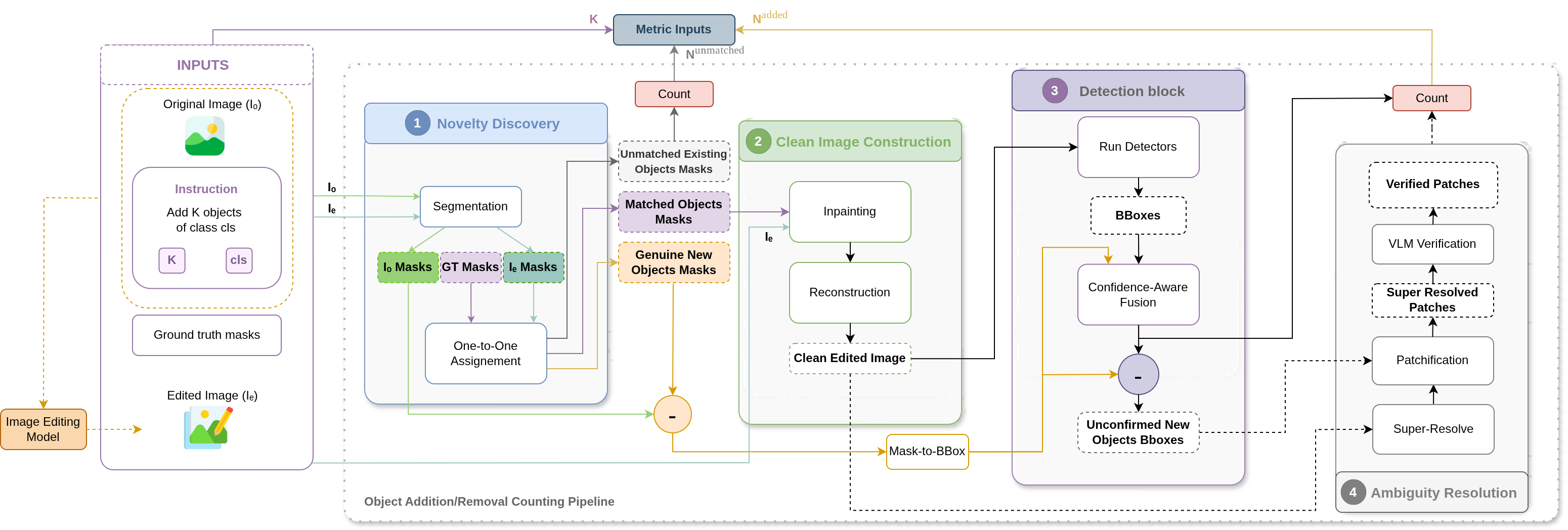}
  \caption{\textbf{`Add'/`Remove' pipeline.} Addition and removal share one
  pipeline (Sec.~\ref{sec:add}); four stages map $I_o$, $I_e$, and the
  panoptic ground truth to the confirmed new instances $N^{\mathrm{added}}$ and
  the unmatched source instances $N^{\mathrm{unmatched}}$: \emph{novelty
  discovery}, \emph{clean-image construction}, dual-detector \emph{detection},
  and VLM \emph{ambiguity resolution}. The two outputs swap roles between
  addition and removal.}
  \label{fig:add}
\end{figure*}

\name{} scores an edit by reconstructing what happened to \emph{every
object instance}, never by asking a model whether the edit succeeded.
Its inputs are a source image $I_o$, an edited image $I_e$, and an
instruction specifying an operation, a target category (or source/target
pair), and a count $K$. From these, \name{} assigns each instance a verdict
(\emph{kept}, \emph{added}, \emph{removed}, or \emph{transformed}), from which
the exact-count and scene-conservation scores follow. The three operations share these components: we
detail them on the \emph{add} pipeline (Fig.~\ref{fig:add}), then show how
\emph{remove} and \emph{change} reuse and extend the same machinery.

\subsection{`Add' pipeline}
\label{sec:add}
The `add' pipeline is where the shared machinery is defined; remove and change
reuse it. From the source image $I_o$, the edited image $I_e$, the panoptic
ground truth, and the instruction ``add $K$ objects of class $\mathrm{cls}$'',
it produces a realized add count, compared against $K$, in four stages
(Fig.~\ref{fig:add}).

\noindent\textbf{Novelty discovery.}
We first gather $\mathrm{cls}$ instances from two grounded sources: the
panoptic ground-truth masks of $I_o$~\cite{coconut}, and presence-gated
concept segmentation (SAM~3~\cite{sam3}) run on \emph{both} $I_o$ and $I_e$
with the category name as prompt, keeping a mask only when its presence score
exceeds $\tau_p$; this gates out hallucinated masks and recovers occluded and
clustered instances, and segmenting the \emph{source} too is what later lets
us repair omissions in the ground truth. A preserved object can move, deform,
or be repainted, so we match the source instances to the edited ones with a
composite \emph{geometric} score robust to this drift. For a source instance
$I_o$ (mask $M_o$) and an edit instance $I_e$ (mask $M_e$),
\begin{equation}
s(I_o,I_e)=\max\!\Big(\max\!\big(\mathrm{IoU},\,c\cdot\mathbf{1}\{c\ge\tau_c\}\big)\cdot\phi,\ \ \sigma\Big),
\label{eq:matchscore}
\end{equation}
with mask overlap $\mathrm{IoU}=|M_o\cap M_e|/|M_o\cup M_e|$, containment
$c=|M_o\cap M_e|/|M_e|$ (for an object the edit split or merged), a size-ratio
factor $\phi=\min(1,|M_o|/|M_e|)$, and a contour term
$\sigma=1/(1+d_{\mathrm{shape}})$ from Hu-moment shape matching, gated to $0$
unless the two centroids lie within a fraction $\tau_d$ of the image diagonal.
Stacking these into a score matrix $S\in\mathbb{R}^{n_o\times n_e}$ whose
entry $S_{ij}$ scores the $i$-th source against the $j$-th edit instance, we
solve the global one-to-one assignment with the Hungarian algorithm,
\begin{equation}
\pi^\star=\arg\max_{\pi}\sum\nolimits_i S_{i,\pi(i)} ,
\label{eq:hungarian}
\end{equation}
keeping a matched pair $(i,\pi(i))$ only when its score $S_{i,\pi(i)}\ge\tau_m$
and its within-intersection pixel similarity stays above $\tau_a$, so a relocated
original is matched to itself while a genuinely new instance is not absorbed
into an existing one. This splits the edited instances into \emph{matched}
instances that already existed and unmatched candidates. Because panoptic
labels are incomplete, each candidate is cross-checked against the SAM-3
segmentation of the \emph{source} image: if a confident counterpart is present
in $I_o$, it is reassigned to \emph{existing} rather than counted as new, and
fragments of one object are merged. The survivors are the \emph{genuine new
instances}. We set $\tau_p\!=\!0.6$, $\tau_c\!=\!0.7$, $\tau_d\!=\!0.3$,
$\tau_m\!=\!0.1$, and $\tau_a\!=\!0.15$. These five thresholds are fixed once
and never retuned; a sweep shows four are inert and even the most sensitive,
$\tau_m$, moves agreement by at most $0.03$
(Table~\ref{tab:thresholds}, App.~\ref{sec:appendix-robustness}).

\noindent\textbf{Clean-image construction.}
To count only what was added, we remove the preserved scene: the matched
(existing) regions are erased from $I_e$ with classical Non-Local Patch-Based
inpainting~\cite{nlpbinpaint}, a non-generative fill that introduces no new
content, and the genuine-new regions are composited back to form a
\emph{clean edited image} that ideally contains the additions alone.
Isolating the new content this way lets a new object be counted even when it
sits in front of or overlaps a pre-existing one.

\noindent\textbf{Detection.}
Two detectors, RF-DETR~\cite{rfdetr} and Co-DETR~\cite{codetr}, run on the
clean image and vote \emph{independently} on the SAM-localized regions: for
each genuine-new mask, each detector is asked only whether a $\mathrm{cls}$ box
overlaps that region, and the mask is confirmed when \emph{both} agree.
Decoupling the vote (each detector agrees with the SAM region, not with the
other detector's exact box) recovers objects that both see but box slightly
differently, while the two-detector gate still blocks single-detector false
positives. Masks that resolve to the same detector box are re-anchored to a
single instance, and a standard cross-detector consensus catches objects SAM
missed entirely.

\noindent\textbf{Ambiguity resolution.}
The few genuine-new candidates the detectors leave unconfirmed are not
discarded blindly. Each is super-resolved
(Real-ESRGAN~\cite{realesrgan}), cropped with context, and passed to a
constrained verifier (Qwen3.6-27B~\cite{qwen36vl}) that answers a single
yes/no question, ``is the boxed object a $\mathrm{cls}$?''. This verifier is invoked
on only $6.4\%$ of instances and only confirms an instance's category on the
rare ambiguous crop; it never counts objects or
judges the edit, which keeps \name{} judge-free and, as the ablation in
Sec.~\ref{sec:ablations} shows, contributes little to the final accuracy.

The pipeline emits two counts (Fig.~\ref{fig:add}): the confirmed new instances
$N^{\mathrm{added}}$ and the unmatched source instances $N^{\mathrm{unmatched}}$.
For addition, $N^{\mathrm{added}}$ is the realized add count $d$ that feeds
count fidelity against $K$, while $N^{\mathrm{unmatched}}$ is the
destroyed-originals count $N_{\mathrm{rem}}$ that drives original-category
conservation (Sec.~\ref{sec:scoring}). Per-sample walkthroughs, including
matching under drift and ground-truth repair, are in App.~\ref{sec:appendix-protocol}
(App.~\ref{sec:addcase}--\ref{sec:matchcase}).

\subsection{`Remove' pipeline}
\label{sec:remove}
Removal is the exact mirror of addition and reuses the four stages of
Sec.~\ref{sec:add} unchanged; only the use of the two outputs swaps
(Fig.~\ref{fig:add}). Given ``remove $K$ objects of class $\mathrm{cls}$'',
novelty discovery matches the source instances to the edited image: a source
instance with a match has \emph{survived} (the deletion failed), while an
unmatched source instance has been \emph{removed}. Here the unmatched-source
output $N^{\mathrm{unmatched}}$ is the realized removed count $r$. The
detection and ambiguity stages now play the opposite role: a genuine-new
instance the dual detectors confirm is not a wanted addition but a
\emph{spurious appearance}, so the detection output $N^{\mathrm{added}}$
becomes the penalty $a$, an object the editor hallucinated while it was
supposed to be deleting. The within-mask appearance test of
Sec.~\ref{sec:add} keeps an object the editor merely faded or left in
place matched, so it is not mistaken for removed. The removed count $r$ and the
spurious additions $a$ feed the score (Sec.~\ref{sec:scoring}).

\subsection{`Change' pipeline}
\label{sec:change}
``Change $K$ objects from class $\mathrm{cls1}$ to class $\mathrm{cls2}$'' is a
removal of $\mathrm{cls1}$ coupled with an addition of $\mathrm{cls2}$, and
counts as success only when the \emph{same} instances change \emph{in place}
(Fig.~\ref{fig:changepipe} of App.~\ref{sec:changecase}). We therefore run the remove pipeline on
$\mathrm{cls1}$ and the add pipeline on $\mathrm{cls2}$ over the same image
pair, each returning its realized count, its conservation penalty, and the
\emph{positions} of the removed $\mathrm{cls1}$ and added $\mathrm{cls2}$
instances. A change is genuine only when a removed $\mathrm{cls1}$ and an added
$\mathrm{cls2}$ coincide in location, so we match the two position sets: the
number of position-coincident pairs is the count of in-place transformations
$t$ ($N^{\mathrm{changed}}$ in the figure), with
$N_{\mathrm{cls1}}^{\mathrm{ip}}$ source removals and
$N_{\mathrm{cls2}}^{\mathrm{ip}}$ target additions accounted for. These
per-branch counts feed the score (Sec.~\ref{sec:scoring}). This
position-aware coupling rejects delete-here-paint-there edits that a naive
remove-plus-add count would accept.

\subsection{Scoring}
\label{sec:scoring}
From the per-instance verdicts, each edit receives a score in $[0,1]$ that
\emph{multiplies} three factors, each catching a failure the others cannot
see,
\begin{equation}
\mathrm{Score} \;=\; \mathrm{CF}\,\cdot\,\mathrm{SC}\,\cdot\,\mathrm{OC},
\label{eq:score}
\end{equation}
where $\mathrm{CF}$ is count fidelity, $\mathrm{SC}$ is scene consistency,
and $\mathrm{OC}$ is original-category conservation. We score exactly
these three because they are what numerical editing requires: CF captures the
requested count, SC and OC that the rest of the image is preserved;
artistic-style fidelity is deliberately out of scope. The product is
unforgiving by design: a model cannot offset a corrupted background by
getting the count right, nor a correct count by overwriting existing
objects, so the score is high only when all three hold. It is also
\emph{necessary}: under a plain mean, a do-nothing editor nearly ties the best
model, whereas the product rejects it (App.~\ref{sec:appendix-robustness}). We give the factors
for the \emph{add} operation; \emph{remove} and \emph{change} reuse the same
template with their own count and conservation terms. Throughout,
$\mathrm{HM}(a,b)=2ab/(a+b)$ is the harmonic mean.

\noindent\textbf{Count fidelity ($\mathrm{CF}$).} Let $K$ be the requested
number of additions and $d$ the number of genuinely new instances the
pipeline confirms (Sec.~\ref{sec:add}). We score the count by the Dice
coefficient (equivalently $F_1$) on counts,
\begin{equation}
\mathrm{CF}(d,K) \;=\; \frac{2\min(d,K)}{d+K}
\;=\; \mathrm{HM}\!\Big(\tfrac{\min(d,K)}{d},\ \tfrac{\min(d,K)}{K}\Big),
\label{eq:cf}
\end{equation}
the harmonic mean of a precision-like term $\min(d,K)/d$ (how many detected
additions were requested) and a recall-like term $\min(d,K)/K$ (how many
requested additions were delivered). Thus $\mathrm{CF}=1$ exactly when
$d=K$, $\mathrm{CF}=0$ when $d=0$, and over- and under-adding are penalized
symmetrically.

\noindent\textbf{Original-category conservation ($\mathrm{OC}$).} Adding
objects must not destroy pre-existing ones of the same category. With
$N_{\mathrm{GT}}$ such instances present before the edit and
$N_{\mathrm{rem}}$ of them flagged as removed or overwritten by the same
instance matching (Sec.~\ref{sec:add}),
\begin{equation}
\mathrm{OC} \;=\; 1-\frac{N_{\mathrm{rem}}}{N_{\mathrm{GT}}}.
\label{eq:oc}
\end{equation}
$\mathrm{OC}=1$ leaves every existing instance intact, $\mathrm{OC}=0$ wipes
them all out. This catches a failure $\mathrm{CF}$ is blind to: reaching $K$
by regenerating or relocating an existing instance instead of genuinely
adding new content.

\noindent\textbf{Scene consistency ($\mathrm{SC}$).} Everything outside the
added region, the background $B$, must be left intact. We first neutralize
the new content by pasting the original pixels back over the added-object
mask $M_{\mathrm{new}}$, giving $I_e^{\mathrm{m}}$, so the
metric measures only the background. $\mathrm{SC}$ is the harmonic mean of a
structural and an appearance term,
\begin{align}
\mathrm{SC}&=\mathrm{HM}\big(\mathrm{SC}_{\mathrm{str}},\,\mathrm{SC}_{\mathrm{app}}\big),\label{eq:sc}\\
\mathrm{SC}_{\mathrm{str}} &= \exp\!\Big(-\!\!\operatorname*{mean}_{(i,j)\in B}
   \big\lVert \mathrm{RAFT}\big(I_o,I_e^{\mathrm{m}}\big)^{(i,j)}\big\rVert_2\Big),
   \label{eq:scstr}\\
\mathrm{SC}_{\mathrm{app}} &= \mathrm{MS\text{-}SSIM}\big(I_o[B],\,I_e^{\mathrm{m}}[B]\big).
   \label{eq:scapp}
\end{align}
$\mathrm{SC}_{\mathrm{str}}$ uses dense optical flow~\cite{raft} to detect
spatial drift: a static background gives zero flow and a score of $1$, while
any shift or warp decays it exponentially. $\mathrm{SC}_{\mathrm{app}}$ uses
multi-scale structural similarity~\cite{msssim} to penalize color,
lighting, and texture corruption. The harmonic mean again requires both to
hold, so a perfectly colored but warped background scores poorly, and vice
versa.

\medskip
\noindent\textbf{Remove and change.} The same product
$\mathrm{Score}=\mathrm{CF}\cdot\mathrm{SC}\cdot\mathrm{OC}$ scores the other
two operations. $\mathrm{SC}$ is unchanged, with the background $B$ taken as
everything outside that operation's edited region; only $\mathrm{CF}$ and
$\mathrm{OC}$ are reinstantiated on the matching verdicts, and we clip
$\mathrm{OC}$ to $[0,1]$.

\noindent\emph{Remove $K$ of class $\mathrm{cls}$.} Mirroring addition, with
$r$ removed and $a$ instances of $\mathrm{cls}$ spuriously added
(Sec.~\ref{sec:remove}),
\begin{equation}
\mathrm{CF}=\frac{2\min(r,K)}{r+K},\qquad
\mathrm{OC}=1-\frac{a}{N_{\mathrm{GT}}}.
\label{eq:rem}
\end{equation}

\noindent\emph{Change $K$ $\mathrm{cls1}$ into $\mathrm{cls2}$.} The unit of
success is the in-place transformation $t$ (Sec.~\ref{sec:change}); $\mathrm{OC}$
penalizes the two ways to fake it, a $\mathrm{cls2}$ painted where no
$\mathrm{cls1}$ was ($f_2\!=\!N_{\mathrm{cls2}}\!-\!N_{\mathrm{cls2}}^{\mathrm{ip}}$)
and a $\mathrm{cls1}$ deleted with no replacement
($f_1\!=\!N_{\mathrm{cls1}}\!-\!N_{\mathrm{cls1}}^{\mathrm{ip}}$):
\begin{equation}
\mathrm{CF}=\frac{2\min(t,K)}{t+K},\qquad
\mathrm{OC}=1-\frac{f_1+f_2}{N_{\mathrm{GT}}^{\mathrm{cls1}}}.
\label{eq:chg}
\end{equation}
We report $\mathrm{Score}$ and its three factors per operation and difficulty
tier; for any set of edits, each factor is averaged over the set and the
$\mathrm{Score}$ is their product (App.~\ref{sec:appendix-leaderboard}).

\section{Experiments}
\label{sec:experiments}
\subsection{Our evaluator matches human counting}
\label{sec:validation}
To show that \name{} is trustworthy, we measure how well it recovers object
counts against a human reference. We assemble a 600-image human-evaluation
set whose edits are produced by Seedream~4.0~\cite{seedream4}, stratified by category (covering 79 of the 80 COCO categories), requested
counts, and the easy, medium, and hard tiers. Its composition is detailed in App.~\ref{sec:hecomp} and the editing prompt in App.~\ref{sec:editprompt}.
\begin{table*}[t]
  \centering
  \footnotesize
  \setlength{\tabcolsep}{3pt}
  \renewcommand{\arraystretch}{1}
  \begin{tabular}{lccccccccccccc}
    \toprule
    & \multicolumn{3}{c}{Add} & \multicolumn{3}{c}{Remove} & \multicolumn{3}{c}{Change} & \multicolumn{4}{c}{Overall ($\mathrm{Score}=\mathrm{CF}\cdot\mathrm{SC}\cdot\mathrm{OC}$)} \\
    \cmidrule(lr){2-4}\cmidrule(lr){5-7}\cmidrule(lr){8-10}\cmidrule(lr){11-14}
    Model & E & M & H & E & M & H & E & M & H & CF & SC & OC & Score \\
    \midrule
    Gemini 2.5 Flash Image~\cite{geminiflash}  & 0.52 & 0.39 & 0.26 & 0.45 & 0.30 & 0.18 & 0.31 & 0.22 & 0.12 & 0.59 & 0.77 & 0.64 & 0.29 \\
    Seedream 4.0~\cite{seedream4}              & \textbf{0.55} & \textbf{0.42} & \textbf{0.28} & \textbf{0.46} & \textbf{0.31} & \textbf{0.20} & \textbf{0.38} & \textbf{0.25} & \textbf{0.14} & \textbf{0.60} & \textbf{0.79} & \textbf{0.66} & \textbf{0.31} \\
    GPT-Image-1~\cite{gptimage1}               & 0.51 & 0.36 & 0.25 & 0.38 & 0.24 & 0.17 & 0.31 & 0.18 & 0.11 & 0.56 & 0.75 & 0.62 & 0.26 \\
    Qwen-Image-Edit~\cite{qwenimage}           & 0.35 & 0.23 & 0.13 & 0.26 & 0.18 & 0.08 & 0.18 & 0.11 & 0.05 & 0.42 & 0.69 & 0.55 & 0.16 \\
    FLUX.1 Kontext~\cite{fluxkontext}          & 0.30 & 0.20 & 0.12 & 0.24 & 0.14 & 0.08 & 0.14 & 0.09 & 0.05 & 0.39 & 0.67 & 0.53 & 0.14 \\
    Step1X-Edit~\cite{step1xedit}              & 0.24 & 0.16 & 0.10 & 0.18 & 0.12 & 0.07 & 0.11 & 0.06 & 0.03 & 0.34 & 0.64 & 0.50 & 0.11 \\
    BAGEL~\cite{bagel}                         & 0.20 & 0.13 & 0.08 & 0.14 & 0.08 & 0.05 & 0.10 & 0.06 & 0.03 & 0.30 & 0.62 & 0.47 & 0.09 \\
    \bottomrule
  \end{tabular}
  \caption{\textbf{Editing-model leaderboard on \name{}.} E/M/H: cell Score
  $\mathrm{CF}_c\cdot\mathrm{SC}_c\cdot\mathrm{OC}_c$ (Eq.~\ref{eq:score}), each
  factor averaged over the edits of that operation$\times$tier cell. Overall:
  CF, SC, OC averaged over all $10{,}944$ edits; Score is their product, not
  the mean of the cell Scores. Bold: best per column; the three proprietary
  models' $95\%$ bootstrap intervals overlap (App.~\ref{sec:appendix-settings}).
  Details: App.~\ref{sec:appendix-leaderboard}.}
  \label{tab:leaderboard}
\end{table*}

Annotating this set by hand is not a convenience but a necessity. Image
editing models still fail to produce the \emph{requested} number of objects,
and the gap widens as the count and scene complexity
grow~\cite{refedit,makeitcount,t2icannotcount}; on our set the number of
target objects actually present in an edited image rarely equals the
prompt's target $K$. The prompt count $K$ is therefore a property of the
\emph{instruction}, not of the image: using it as the reference would
conflate the editor's failure with the evaluator's, and reward an evaluator
for parroting $K$ rather than for counting what is on the canvas. We instead
annotate the true number of target-category objects in each edited image and
validate every evaluator against this \emph{human} count, never against $K$.

We compare \name{} against two single-tool counters on this set: a
vision-language model used as a solo counter (Qwen3.6-27B~\cite{qwen36vl} in
thinking mode) and a single open-vocabulary detector
(Grounding-DINO~\cite{groundingdino}). Both are deliberate, conservative
choices: Qwen3.6-27B is a top recent VLM and the \emph{same} model \name{}
uses internally for the rare crop check, so this head-to-head isolates that the
VLM is not the source of our accuracy (confirmed by the ablation in
Table~\ref{tab:ablation}); and Grounding-DINO is precisely the single
open-vocabulary detector EdiVal-Agent~\cite{edival} uses to score count-change
edits (full rationale in App.~\ref{sec:appendix-crosseditor}).

Each evaluator estimates the number of target objects in the edited image,
and we report how often that estimate matches the human count, per
difficulty tier (Table~\ref{tab:agreement}; visualized in
Fig.~\ref{fig:validation} of App.~\ref{sec:agreement-fig}).
\name{} tracks the human count far more closely than either alternative at
every tier (0.89 on easy down to 0.76 on hard), though it too declines on the
busiest scenes; the solo VLM degrades sharply with count and complexity; and
the single detector is the \emph{least} reliable of all: on edited images
Grounding-DINO routinely
over- and under-counts through duplicate, merged, or hallucinated boxes, and
falls below even the VLM. Its poor reliability here is the point: a single
open-vocabulary detector is not a trustworthy mechanism for the numerical
problem, which motivates the dual-detector consensus at the core of \name{}. The same ordering holds as the object count grows
(Fig.~\ref{fig:valcount} of App.~\ref{sec:agreement-fig}).

\begin{table}[t]
  \centering
  \small
  \setlength{\tabcolsep}{5pt}
  \begin{tabular}{lcccc}
    \toprule
    Evaluator & Easy & Medium & Hard & Overall \\
    \midrule
    Grounding-DINO~\cite{groundingdino} & 0.37 & 0.26 & 0.23 & 0.28 \\
    Qwen3.6-27B (thinking)                     & 0.71 & 0.54 & 0.45 & 0.55 \\
    \textbf{\name{} (ours)}                                & \textbf{0.89} & \textbf{0.82} & \textbf{0.76} & \textbf{0.82} \\
    \bottomrule
  \end{tabular}
  \caption{\textbf{Exact-count agreement with the human count}
  ($\uparrow$), per difficulty tier, on the 600-image study.}
  \label{tab:agreement}
\end{table}

\noindent\textbf{Validating the full Score, not counting alone.} We also
validate every other factor of Eq.~\ref{eq:score} against humans:
original-category conservation $0.97$ (balanced accuracy $0.90$),
change-correspondence $0.95$, scene consistency (\name{} SC versus human scene-consistency rating) $\rho=0.87$, and the combined
Score (\name{} Score versus human overall rating) $\rho=0.88$ (App.~\ref{sec:appendix-fullscore}); and \name{} stays within
$\sim\!0.10$ of a two-annotator human ceiling, a gap flat across tiers
(App.~\ref{sec:appendix-annotation}).

\subsection{\name{} stays reliable across editors}
\label{sec:crosseditor}
The validation in Sec.~\ref{sec:validation} uses one editor. To show it is not tuned to one model, we re-run it on all
seven editors over the same $200$ source images (a paired design). As the editors
weaken from Seedream to BAGEL, \name{}'s exact-count agreement barely moves
($0.83$ to $0.79$, std $0.013$), while the single-tool counters collapse far
faster: the solo VLM falls $0.54$ to $0.37$ and the single detector $0.25$ to
$0.16$ (about $4\times$ and $2\times$ wider; Table~\ref{tab:cross}), and the
\name{} ranking still matches humans (Spearman $\rho=0.96$).

\begin{table}[t]
  \centering\footnotesize
  \setlength{\tabcolsep}{5pt}
  \begin{tabular}{lccc}
    \toprule
    Editor & \name{} & Solo VLM & Single det. \\
    \midrule
    Seedream 4.0~\cite{seedream4}      & \textbf{0.83} & 0.54 & 0.25 \\
    Gemini 2.5~\cite{geminiflash}      & 0.81 & 0.53 & 0.24 \\
    GPT-Image-1~\cite{gptimage1}       & 0.81 & 0.51 & 0.23 \\
    Qwen-Image-Edit~\cite{qwenimage}   & 0.80 & 0.47 & 0.22 \\
    FLUX.1 Kontext~\cite{fluxkontext}  & 0.80 & 0.43 & 0.20 \\
    Step1X-Edit~\cite{step1xedit}      & 0.80 & 0.41 & 0.18 \\
    BAGEL~\cite{bagel}                 & 0.79 & 0.37 & 0.16 \\
    \midrule
    Mean (std) & \textbf{0.81}\,(0.013) & 0.47\,(0.065) & 0.21\,(0.033) \\
    \bottomrule
  \end{tabular}
  \caption{\textbf{\name{} is stable across editors.} Exact-count agreement
  with the human count ($200$ edited images per editor, drawn from the same source images).}
  \label{tab:cross}
\end{table}

\subsection{State-of-the-art editors fail at numerical editing}
\label{sec:leaderboard}
Table~\ref{tab:leaderboard} reports the main result: leading editors miss the
requested count on most edits, especially for \emph{remove} and \emph{change}
and in cluttered scenes. Decomposing the score exposes two
compounding failures: count fidelity (CF) is the weakest factor, and
original-category conservation (OC) is also low, as editors frequently
overwrite or delete existing same-category objects when adding, and
spuriously insert new ones when removing.
Only scene consistency (SC) stays high, so the surrounding background is
preserved even when the target objects are not. The proprietary systems lead
the open models by a clear margin, yet even the strongest, Seedream~4.0,
reaches only $0.31$ overall. Because the Score multiplies three factors,
even the best system fails at least one of count, conservation, or scene
preservation on most edits.
\subsection{Ablations}
\label{sec:ablations}
We ablate each component on the 600 human-annotated images of
Sec.~\ref{sec:validation}, removing one at a time and re-measuring exact-count
agreement against the human count
(Table~\ref{tab:ablation} of App.~\ref{sec:appendix-ablation}). Each part is load-bearing, but to different
degrees. Removing the matcher's one-to-one alignment is the most damaging
(0.82 to 0.36): instead of re-matching every source instance to its
drifted counterpart in the edit, the source mask is reused \emph{in place},
and the editor's small shifts in position and size leave it covering nothing,
so the preserved object is never masked out and is recounted as new. The next
three each cost about ten points: clean-image isolation inpaints the matched
objects out, and without it the detectors count pre-existing and added
instances together; ground-truth completion guards the opposite error,
rescuing a real object the incomplete panoptic labels would otherwise score as
new; and a single detector (averaged over keeping only RF-DETR or only
Co-DETR) admits class false positives that the dual-detector agreement filters.
The constrained VLM verifier matters least (0.82 to 0.79): it only
confirms a category on the rare ambiguous crop, so \name{}'s accuracy rests on
grounded geometry and detection, not the VLM, and the protocol stays
judge-free.

\section{Conclusion}
\label{sec:conclusion}

We argued that numerical image editing must be \emph{verified}, not judged,
and proposed \name{}, which combines three components: a numerically
controlled benchmark with instance-level ground truth, a judge-free evaluation
protocol, and a scoring metric that jointly rewards the exact requested count
and the conservation of the rest of the scene. \name{} matches human counting
where VLMs and single detectors collapse, and benchmarking leading editors
reveals a failure the field had overlooked: even the strongest systems still cannot count, delivering the wrong number of
objects on most edits, or reaching the target only by overwriting what was
already there.

\medskip
\noindent\textbf{Scope and limitations.} \name{} is scoped to the $80$ COCO
thing-categories by design: reliable judge-free confirmation needs a fixed
detector vocabulary, and outside it, the closed-set detectors return no
detections. Extending to rarer categories needs open-vocabulary detectors,
themselves less reliable for counting (the failure \name{} avoids). Open-vocabulary
numerical editing is the natural next step. Extending it to other domains
and detectors (e.g., aerial imagery) is a further direction. \name{} also scores the count and
scene conservation, not which specific instances are chosen: any valid
$K$-subset satisfies the instruction when $K\!<\!N$, while conservation still
blocks reaching the count by replacing existing ones. Numerical edits with attribute and positioning constraints, which instances to edit and with which attributes, are the next research axes.

\noindent\textbf{Acknowledgments.} Funded by the European Union. Views and opinions expressed are however those of the author(s) only and do not necessarily reflect those of the European Union nor the European Commission. Neither the European Union nor the granting authority can be held responsible for them. This work was supported under the Horizon Europe Project CustomAI (grant number 101226029) and the FactoryIA supercomputer (funded by the Ile-de-France Regional Council).

{
    \small
    \bibliographystyle{ieeenat_fullname}
    \bibliography{main}

\begin{thebibliography}{47}
\providecommand{\natexlab}[1]{#1}
\providecommand{\url}[1]{\texttt{#1}}
\expandafter\ifx\csname urlstyle\endcsname\relax
  \providecommand{\doi}[1]{doi: #1}\else
  \providecommand{\doi}{doi: \begingroup \urlstyle{rm}\Url}\fi

\bibitem[Basu et~al.(2023)Basu, Saberi, Bhardwaj, Chegini, Massiceti, Sanjabi, Hu, and Feizi]{editval2023}
Samyadeep Basu, Mehrdad Saberi, Shweta Bhardwaj, Atoosa~Malemir Chegini, Daniela Massiceti, Maziar Sanjabi, Shell~Xu Hu, and Soheil Feizi.
\newblock {EditVal}: Benchmarking diffusion based text-guided image editing methods.
\newblock \emph{arXiv preprint arXiv:2310.02426}, 2023.

\bibitem[Bertalmio et~al.(2001)Bertalmio, Bertozzi, and Sapiro]{navierstokesinpaint}
M. Bertalmio, A.L. Bertozzi, and G. Sapiro.
\newblock Navier-stokes, fluid dynamics, and image and video inpainting.
\newblock In \emph{Proceedings of the 2001 IEEE Computer Society Conference on Computer Vision and Pattern Recognition. CVPR 2001}, pages I--I, 2001.

\bibitem[Binyamin et~al.(2024)Binyamin, Tewel, Segev, Hirsch, Rassin, and Chechik]{makeitcount}
Lital Binyamin, Yoad Tewel, Hilit Segev, Eran Hirsch, Royi Rassin, and Gal Chechik.
\newblock Make it count: Text-to-image generation with an accurate number of objects.
\newblock \emph{arXiv preprint arXiv:2406.10210}, 2024.

\bibitem[{Black Forest Labs} et~al.(2025){Black Forest Labs}, Batifol, Blattmann, Boesel, Consul, Diagne, Dockhorn, English, English, Esser, Kulal, Lacey, Levi, Li, Lorenz, M{\"u}ller, Podell, Rombach, Saini, Sauer, and Smith]{fluxkontext}
{Black Forest Labs}, Stephen Batifol, Andreas Blattmann, Frederic Boesel, Saksham Consul, Cyril Diagne, Tim Dockhorn, Jack English, Zion English, Patrick Esser, Sumith Kulal, Kyle Lacey, Yam Levi, Cheng Li, Dominik Lorenz, Jonas M{\"u}ller, Dustin Podell, Robin Rombach, Harry Saini, Axel Sauer, and Luke Smith.
\newblock {FLUX.1} {Kontext}: Flow matching for in-context image generation and editing in latent space.
\newblock \emph{arXiv preprint arXiv:2506.15742}, 2025.

\bibitem[Brooks et~al.(2022)Brooks, Holynski, and Efros]{instructpix2pix}
Tim Brooks, Aleksander Holynski, and Alexei~A. Efros.
\newblock {InstructPix2Pix}: Learning to follow image editing instructions.
\newblock \emph{arXiv preprint arXiv:2211.09800}, 2022.

\bibitem[Carion et~al.(2025)Carion, Gustafson, Hu, Debnath, Hu, Suris, Ryali, Alwala, Khedr, Huang, Lei, Ma, Guo, Kalla, Marks, Greer, Wang, Sun, R{\"a}dle, Afouras, Mavroudi, Xu, Wu, Zhou, Momeni, Hazra, Ding, Vaze, Porcher, Li, Li, Kamath, Cheng, Doll{\'a}r, Ravi, Saenko, Zhang, and Feichtenhofer]{sam3}
Nicolas Carion, Laura Gustafson, Yuan-Ting Hu, Shoubhik Debnath, Ronghang Hu, Didac Suris, Chaitanya Ryali, Kalyan~Vasudev Alwala, Haitham Khedr, Andrew Huang, Jie Lei, Tengyu Ma, Baishan Guo, Arpit Kalla, Markus Marks, Joseph Greer, Meng Wang, Peize Sun, Roman R{\"a}dle, Triantafyllos Afouras, Effrosyni Mavroudi, Katherine Xu, Tsung-Han Wu, Yu Zhou, Liliane Momeni, Rishi Hazra, Shuangrui Ding, Sagar Vaze, Francois Porcher, Feng Li, Siyuan Li, Aishwarya Kamath, Ho~Kei Cheng, Piotr Doll{\'a}r, Nikhila Ravi, Kate Saenko, Pengchuan Zhang, and Christoph Feichtenhofer.
\newblock {SAM 3}: Segment anything with concepts.
\newblock \emph{arXiv preprint arXiv:2511.16719}, 2025.

\bibitem[Caron et~al.(2021)Caron, Touvron, Misra, J{\'e}gou, Mairal, Bojanowski, and Joulin]{dino}
Mathilde Caron, Hugo Touvron, Ishan Misra, Herv{\'e} J{\'e}gou, Julien Mairal, Piotr Bojanowski, and Armand Joulin.
\newblock Emerging properties in self-supervised vision transformers.
\newblock \emph{arXiv preprint arXiv:2104.14294}, 2021.

\bibitem[Chen et~al.(2025{\natexlab{a}})Chen, Wei, Zhou, and Kumar]{adiee}
Sherry~X. Chen, Yi Wei, Luowei Zhou, and Suren Kumar.
\newblock {ADIEE}: Automatic dataset creation and scorer for instruction-guided image editing evaluation.
\newblock \emph{arXiv preprint arXiv:2507.07317}, 2025{\natexlab{a}}.

\bibitem[Chen et~al.(2025{\natexlab{b}})Chen, Zhang, Zhang, Yu, Wang, Wang, Lin, Wang, Yang, Li, Lin, Xie, Leong, Wang, Wu, and Zhou]{edival}
Tianyu Chen, Yasi Zhang, Zhi Zhang, Peiyu Yu, Shu Wang, Zhendong Wang, Kevin Lin, Xiaofei Wang, Zhengyuan Yang, Linjie Li, Chung-Ching Lin, Jianwen Xie, Oscar Leong, Lijuan Wang, Ying~Nian Wu, and Mingyuan Zhou.
\newblock {EdiVal-Agent}: An object-centric framework for automated, fine-grained evaluation of multi-turn editing.
\newblock \emph{arXiv preprint arXiv:2509.13399}, 2025{\natexlab{b}}.

\bibitem[Deng et~al.(2025)Deng, Zhu, Li, Gou, Li, Wang, Zhong, Yu, Nie, Song, Shi, and Fan]{bagel}
Chaorui Deng, Deyao Zhu, Kunchang Li, Chenhui Gou, Feng Li, Zeyu Wang, Shu Zhong, Weihao Yu, Xiaonan Nie, Ziang Song, Guang Shi, and Haoqi Fan.
\newblock Emerging properties in unified multimodal pretraining.
\newblock \emph{arXiv preprint arXiv:2505.14683}, 2025.

\bibitem[Deng et~al.(2024)Deng, Yu, Wang, Shen, and Chen]{coconut}
Xueqing Deng, Qihang Yu, Peng Wang, Xiaohui Shen, and Liang-Chieh Chen.
\newblock {COCONut}: Modernizing {COCO} segmentation.
\newblock \emph{arXiv preprint arXiv:2404.08639}, 2024.

\bibitem[{Google}(2025)]{geminiflash}
{Google}.
\newblock Introducing {Gemini 2.5 Flash Image}, our state-of-the-art image model.
\newblock {\Urlmuskip=0mu\relax\url{https://developers.googleblog.com/en/introducing-gemini-2-5-flash-image/}}, 2025.
\newblock Gemini 2.5 Flash Image, also known as Nano Banana.

\bibitem[Guo et~al.(2025{\natexlab{a}})Guo, Huang, Shi, Song, and Zhang]{vlmcount20}
Xuyang Guo, Zekai Huang, Zhenmei Shi, Zhao Song, and Jiahao Zhang.
\newblock Your vision-language model can't even count to 20: Exposing the failures of {VLMs} in compositional counting.
\newblock \emph{arXiv preprint arXiv:2510.04401}, 2025{\natexlab{a}}.

\bibitem[Guo et~al.(2025{\natexlab{b}})Guo, Huo, Liang, Shi, Song, Zhang, and Zhuang]{t2icannotcount}
Xuyang Guo, Jiayan Huo, Yingyu Liang, Zhenmei Shi, Zhao Song, Jiahao Zhang, and Zhen Zhuang.
\newblock Text-to-image diffusion models cannot count, and prompt refinement cannot help.
\newblock \emph{arXiv preprint arXiv:2503.06884}, 2025{\natexlab{b}}.

\bibitem[Heusel et~al.(2017)Heusel, Ramsauer, Unterthiner, Nessler, and Hochreiter]{fid}
Martin Heusel, Hubert Ramsauer, Thomas Unterthiner, Bernhard Nessler, and Sepp Hochreiter.
\newblock {GAN}s trained by a two time-scale update rule converge to a local {N}ash equilibrium.
\newblock \emph{arXiv preprint arXiv:1706.08500}, 2017.

\bibitem[Huang et~al.(2025)Huang, Zhang, Jia, Zheng, Gao, Park, Han, and Krishna]{soc}
Weikai Huang, Jieyu Zhang, Taoyang Jia, Chenhao Zheng, Ziqi Gao, Jae~Sung Park, Winson Han, and Ranjay Krishna.
\newblock Synthetic object compositions for scalable and accurate learning in detection, segmentation, and grounding.
\newblock \emph{arXiv preprint arXiv:2510.09110}, 2025.

\bibitem[Jia et~al.(2025)Jia, Huang, Tang, Qiao, Liao, Cao, Zhao, Feng, Gu, Yin, Bai, Ouyang, Chen, Zhao, Hu, Wang, Xie, and Lin]{compbench}
Bohan Jia, Wenxuan Huang, Yuntian Tang, Junbo Qiao, Jincheng Liao, Shaosheng Cao, Fei Zhao, Zhaopeng Feng, Zhouhong Gu, Zhenfei Yin, Lei Bai, Wanli Ouyang, Lin Chen, Fei Zhao, Yao Hu, Zihan Wang, Yuan Xie, and Shaohui Lin.
\newblock {CompBench}: Benchmarking complex instruction-guided image editing.
\newblock \emph{arXiv preprint arXiv:2505.12200}, 2025.

\bibitem[Jocher et~al.(2026)Jocher, Qiu, Liu, Lyu, Akyon, and Kalfaoglu]{jocher2026ultralyticsyolo26unifiedrealtime}
Glenn Jocher, Jing Qiu, Mengyu Liu, Shuai Lyu, Fatih~Cagatay Akyon, and Muhammet~Esat Kalfaoglu.
\newblock Ultralytics {YOLO26}: Unified real-time end-to-end vision models.
\newblock \emph{arXiv preprint arXiv:2606.03748}, 2026.

\bibitem[Ku et~al.(2023)Ku, Jiang, Wei, Yue, and Chen]{viescore}
Max Ku, Dongfu Jiang, Cong Wei, Xiang Yue, and Wenhu Chen.
\newblock {VIEScore}: Towards explainable metrics for conditional image synthesis evaluation.
\newblock \emph{arXiv preprint arXiv:2312.14867}, 2023.

\bibitem[Kumar et~al.(2026)Kumar, Tayebati, Naik, Krishnan, and Trivedi]{vlmjudgescore}
Divake Kumar, Sina Tayebati, Devashri Naik, Ranganath Krishnan, and Amit~Ranjan Trivedi.
\newblock {VLM} judges can rank but cannot score: Task-dependent uncertainty in multimodal evaluation.
\newblock \emph{arXiv preprint arXiv:2604.25235}, 2026.

\bibitem[Lin et~al.(2014)Lin, Maire, Belongie, Bourdev, Girshick, Hays, Perona, Ramanan, Zitnick, and Doll{\'a}r]{coco}
Tsung-Yi Lin, Michael Maire, Serge Belongie, Lubomir Bourdev, Ross Girshick, James Hays, Pietro Perona, Deva Ramanan, C.~Lawrence Zitnick, and Piotr Doll{\'a}r.
\newblock Microsoft {COCO}: Common objects in context.
\newblock \emph{arXiv preprint arXiv:1405.0312}, 2014.

\bibitem[Liu et~al.(2026)Liu, Weingord, Mittal, Dungarwal, Nandula, Ni, Basu, Chen, Ahmed, Li, Zhang, Goswami, Mukherjee, Kveton, Mathur, Dernoncourt, Zhao, Wang, Rossi, Tu, and Du]{mllmjudge}
Runzhou Liu, Hailey Weingord, Sejal Mittal, Prakhar Dungarwal, Anusha Nandula, Bo Ni, Samyadeep Basu, Hongjie Chen, Nesreen~K. Ahmed, Li Li, Jiayi Zhang, Koustava Goswami, Subhojyoti Mukherjee, Branislav Kveton, Puneet Mathur, Franck Dernoncourt, Yue Zhao, Yu Wang, Ryan~A. Rossi, Zhengzhong Tu, and Hongru Du.
\newblock Human-aligned {MLLM} judges for fine-grained image editing evaluation: A benchmark, framework, and analysis.
\newblock \emph{arXiv preprint arXiv:2602.13028}, 2026.

\bibitem[Liu et~al.(2023)Liu, Zeng, Ren, Li, Zhang, Yang, Jiang, Li, Yang, Su, Zhu, and Zhang]{groundingdino}
Shilong Liu, Zhaoyang Zeng, Tianhe Ren, Feng Li, Hao Zhang, Jie Yang, Qing Jiang, Chunyuan Li, Jianwei Yang, Hang Su, Jun Zhu, and Lei Zhang.
\newblock Grounding {DINO}: Marrying {DINO} with grounded pre-training for open-set object detection.
\newblock \emph{arXiv preprint arXiv:2303.05499}, 2023.

\bibitem[Liu et~al.(2025)Liu, Han, Xing, Yin, Wang, Cheng, Liao, Wang, Fu, Han, Li, Peng, Sun, Wu, Cai, Ge, Ming, Xia, Zeng, Zhu, Jiao, Zhang, Yu, and Jiang]{step1xedit}
Shiyu Liu, Yucheng Han, Peng Xing, Fukun Yin, Rui Wang, Wei Cheng, Jiaqi Liao, Yingming Wang, Honghao Fu, Chunrui Han, Guopeng Li, Yuang Peng, Quan Sun, Jingwei Wu, Yan Cai, Zheng Ge, Ranchen Ming, Lei Xia, Xianfang Zeng, Yibo Zhu, Binxing Jiao, Xiangyu Zhang, Gang Yu, and Daxin Jiang.
\newblock {Step1X-Edit}: A practical framework for general image editing.
\newblock \emph{arXiv preprint arXiv:2504.17761}, 2025.

\bibitem[Ma et~al.(2024)Ma, Ji, Ye, Lin, Wang, Zheng, Zhou, Sun, and Ji]{i2ebench}
Yiwei Ma, Jiayi Ji, Ke Ye, Weihuang Lin, Zhibin Wang, Yonghan Zheng, Qiang Zhou, Xiaoshuai Sun, and Rongrong Ji.
\newblock {I2EBench}: A comprehensive benchmark for instruction-based image editing.
\newblock \emph{arXiv preprint arXiv:2408.14180}, 2024.

\bibitem[Newson et~al.(2017)Newson, Almansa, Gousseau, and P{\'e}rez]{nlpbinpaint}
Alasdair Newson, Andr{\'e}s Almansa, Yann Gousseau, and Patrick P{\'e}rez.
\newblock Non-local patch-based image inpainting.
\newblock \emph{Image Processing On Line}, 7:\penalty0 373--385, 2017.

\bibitem[{OpenAI}(2025)]{gptimage1}
{OpenAI}.
\newblock {Gpt-image-1}: {OpenAI}'s multimodal image generation model.
\newblock {\Urlmuskip=0mu\relax\url{https://platform.openai.com/docs/models/gpt-image-1}}, 2025.
\newblock Accessed: 2026-06-22.

\bibitem[Pan et~al.(2025)Pan, He, Mao, Han, Jiang, Zhang, and Liu]{icebench}
Yulin Pan, Xiangteng He, Chaojie Mao, Zhen Han, Zeyinzi Jiang, Jingfeng Zhang, and Yu Liu.
\newblock {ICE-Bench}: A unified and comprehensive benchmark for image creating and editing.
\newblock \emph{arXiv preprint arXiv:2503.14482}, 2025.

\bibitem[Park et~al.(2026)Park, Choi, Kang, Lee, Shin, and Shim]{mllmperceptualbias}
Seojeong Park, Jiho Choi, Junyong Kang, Seonho Lee, Jaeyo Shin, and Hyunjung Shim.
\newblock Mitigating perceptual judgment bias in multimodal {LLM}-as-a-judge via perceptual perturbation and reward modeling.
\newblock \emph{arXiv preprint arXiv:2606.02578}, 2026.

\bibitem[Pathiraja et~al.(2025)Pathiraja, Patel, Singh, Yang, and Baral]{refedit}
Bimsara Pathiraja, Maitreya Patel, Shivam Singh, Yezhou Yang, and Chitta Baral.
\newblock {RefEdit}: A benchmark and method for improving instruction-based image editing model on referring expressions.
\newblock \emph{arXiv preprint arXiv:2506.03448}, 2025.

\bibitem[Qian et~al.(2025)Qian, Lu, Fu, Wang, Chen, Yang, Hu, and Gan]{giebench}
Yusu Qian, Jiasen Lu, Tsu-Jui Fu, Xinze Wang, Chen Chen, Yinfei Yang, Wenze Hu, and Zhe Gan.
\newblock {GIE-Bench}: Towards grounded evaluation for text-guided image editing.
\newblock \emph{arXiv preprint arXiv:2505.11493}, 2025.

\bibitem[{Qwen Team}(2026)]{qwen36vl}
{Qwen Team}.
\newblock {Qwen3.6-27B}: Flagship-level coding in a {27B} dense model, 2026.

\bibitem[Radford et~al.(2021)Radford, Kim, Hallacy, Ramesh, Goh, Agarwal, Sastry, Askell, Mishkin, Clark, Krueger, and Sutskever]{clip}
Alec Radford, Jong~Wook Kim, Chris Hallacy, Aditya Ramesh, Gabriel Goh, Sandhini Agarwal, Girish Sastry, Amanda Askell, Pamela Mishkin, Jack Clark, Gretchen Krueger, and Ilya Sutskever.
\newblock Learning transferable visual models from natural language supervision.
\newblock \emph{arXiv preprint arXiv:2103.00020}, 2021.

\bibitem[Robinson et~al.(2025)Robinson, Robicheaux, Popov, Ramanan, and Peri]{rfdetr}
Isaac Robinson, Peter Robicheaux, Matvei Popov, Deva Ramanan, and Neehar Peri.
\newblock {RF}-{DETR}: Neural architecture search for real-time detection transformers.
\newblock \emph{arXiv preprint arXiv:2511.09554}, 2025.

\bibitem[Sengupta et~al.(2025)Sengupta, Moradinasab, Liu, and Brown]{vlmcountsynthetic}
Saurav Sengupta, Nazanin Moradinasab, Jiebei Liu, and Donald~E. Brown.
\newblock Can vision-language models count? {A} synthetic benchmark and analysis of attention-based interventions.
\newblock \emph{arXiv preprint arXiv:2511.17722}, 2025.

\bibitem[{Team Seedream} et~al.(2025){Team Seedream}, Chen, Gao, Gong, Guo, Guo, Guo, Hou, Huang, Huang, Jian, Kuang, Lai, Li, Li, Lian, Liao, Liu, Liu, Lu, Luo, Ou, Shi, Shi, Sun, Tian, Tian, Wang, Wang, Wang, Wang, Wu, Wu, Wu, Wu, Xia, Xiao, Xu, Yan, Yang, Yang, Zhai, Zhang, Zhang, Zhang, Zhang, Zhang, Zhao, Zhao, and Zhu]{seedream4}
{Team Seedream}, Yunpeng Chen, Yu Gao, Lixue Gong, Meng Guo, Qiushan Guo, Zhiyao Guo, Xiaoxia Hou, Weilin Huang, Yixuan Huang, Xiaowen Jian, Huafeng Kuang, Zhichao Lai, Fanshi Li, Liang Li, Xiaochen Lian, Chao Liao, Liyang Liu, Wei Liu, Yanzuo Lu, Zhengxiong Luo, Tongtong Ou, Guang Shi, Yichun Shi, Shiqi Sun, Yu Tian, Zhi Tian, Peng Wang, Rui Wang, Xun Wang, Ye Wang, Guofeng Wu, Jie Wu, Wenxu Wu, Yonghui Wu, Xin Xia, Xuefeng Xiao, Shuang Xu, Xin Yan, Ceyuan Yang, Jianchao Yang, Zhonghua Zhai, Chenlin Zhang, Heng Zhang, Qi Zhang, Xinyu Zhang, Yuwei Zhang, Shijia Zhao, Wenliang Zhao, and Wenjia Zhu.
\newblock Seedream 4.0: Toward next-generation multimodal image generation.
\newblock \emph{arXiv preprint arXiv:2509.20427}, 2025.

\bibitem[Teed and Deng(2020)]{raft}
Zachary Teed and Jia Deng.
\newblock {RAFT}: Recurrent all-pairs field transforms for optical flow.
\newblock \emph{arXiv preprint arXiv:2003.12039}, 2020.

\bibitem[Wang et~al.(2021)Wang, Xie, Dong, and Shan]{realesrgan}
Xintao Wang, Liangbin Xie, Chao Dong, and Ying Shan.
\newblock {Real-ESRGAN}: Training real-world blind super-resolution with pure synthetic data.
\newblock \emph{arXiv preprint arXiv:2107.10833}, 2021.

\bibitem[Wang et~al.(2003)Wang, Simoncelli, and Bovik]{msssim}
Z. Wang, Eero Simoncelli, and Alan Bovik.
\newblock Multiscale structural similarity for image quality assessment.
\newblock pages 1398 -- 1402 Vol.2, 2003.

\bibitem[Wu et~al.(2025{\natexlab{a}})Wu, Li, Zhou, Lin, Gao, Yan, Yin, Bai, Xu, Chen, Chen, Tang, Zhang, Wang, Yang, Yu, Cheng, Liu, Li, Zhang, Meng, Wei, Ni, Chen, Cao, Peng, Qu, Wu, Wang, Yu, Wen, Feng, Xu, Wang, Zhang, Zhu, Wu, Cai, and Liu]{qwenimage}
Chenfei Wu, Jiahao Li, Jingren Zhou, Junyang Lin, Kaiyuan Gao, Kun Yan, Sheng-ming Yin, Shuai Bai, Xiao Xu, Yilei Chen, Yuxiang Chen, Zecheng Tang, Zekai Zhang, Zhengyi Wang, An Yang, Bowen Yu, Chen Cheng, Dayiheng Liu, Deqing Li, Hang Zhang, Hao Meng, Hu Wei, Jingyuan Ni, Kai Chen, Kuan Cao, Liang Peng, Lin Qu, Minggang Wu, Peng Wang, Shuting Yu, Tingkun Wen, Wensen Feng, Xiaoxiao Xu, Yi Wang, Yichang Zhang, Yongqiang Zhu, Yujia Wu, Yuxuan Cai, and Zenan Liu.
\newblock {Qwen-Image} technical report.
\newblock \emph{arXiv preprint arXiv:2508.02324}, 2025{\natexlab{a}}.

\bibitem[Wu et~al.(2025{\natexlab{b}})Wu, Li, Hu, Ye, Zeng, Yu, Zhu, Schiele, Yang, and Yang]{krisbench}
Yongliang Wu, Zonghui Li, Xinting Hu, Xinyu Ye, Xianfang Zeng, Gang Yu, Wenbo Zhu, Bernt Schiele, Ming-Hsuan Yang, and Xu Yang.
\newblock {KRIS-Bench}: Benchmarking next-level intelligent image editing models.
\newblock \emph{arXiv preprint arXiv:2505.16707}, 2025{\natexlab{b}}.

\bibitem[Ye et~al.(2025)Ye, He, Li, Lin, Yuan, Yan, Hou, and Yuan]{imgedit}
Yang Ye, Xianyi He, Zongjian Li, Bin Lin, Shenghai Yuan, Zhiyuan Yan, Bohan Hou, and Li Yuan.
\newblock {ImgEdit}: A unified image editing dataset and benchmark.
\newblock \emph{arXiv preprint arXiv:2505.20275}, 2025.

\bibitem[Zhang et~al.(2022)Zhang, Li, Liu, Zhang, Su, Zhu, Ni, and Shum]{zhang2022dino}
Hao Zhang, Feng Li, Shilong Liu, Lei Zhang, Hang Su, Jun Zhu, Lionel~M. Ni, and Heung-Yeung Shum.
\newblock {DINO}: {DETR} with improved denoising anchor boxes for end-to-end object detection.
\newblock \emph{arXiv preprint arXiv:2203.03605}, 2022.

\bibitem[Zhang et~al.(2026)Zhang, Bai, Li, Liang, Tian, Li, An, Zhang, Korhonen, Zhang, Wang, and Tan]{vibe}
Huanyu Zhang, Xuehai Bai, Chengzu Li, Chen Liang, Haochen Tian, Haodong Li, Ruichuan An, Yifan Zhang, Anna Korhonen, Zhang Zhang, Liang Wang, and Tieniu Tan.
\newblock How well do models follow visual instructions? {VIBE}: A systematic benchmark for visual instruction-driven image editing.
\newblock \emph{arXiv preprint arXiv:2602.01851}, 2026.

\bibitem[Zhang et~al.(2023)Zhang, Mo, Chen, Sun, and Su]{magicbrush}
Kai Zhang, Lingbo Mo, Wenhu Chen, Huan Sun, and Yu Su.
\newblock {MagicBrush}: A manually annotated dataset for instruction-guided image editing.
\newblock \emph{arXiv preprint arXiv:2306.10012}, 2023.

\bibitem[Zhou et~al.(2024)Zhou, Guo, Wang, Chang, and Wu]{augsurvey}
Yue Zhou, Chenlu Guo, Xu Wang, Yi Chang, and Yuan Wu.
\newblock A survey on data augmentation in large model era.
\newblock \emph{arXiv preprint arXiv:2401.15422}, 2024.

\bibitem[Zong et~al.(2022)Zong, Song, and Liu]{codetr}
Zhuofan Zong, Guanglu Song, and Yu Liu.
\newblock {DETRs} with collaborative hybrid assignments training.
\newblock \emph{arXiv preprint arXiv:2211.12860}, 2022.

\end{thebibliography}
}

\clearpage
\appendix
\clearpage
\maketitlesupplementary
\label{sec:supp}

\setcounter{figure}{0}
\renewcommand{\thefigure}{S\arabic{figure}}
\setcounter{table}{0}
\renewcommand{\thetable}{S\arabic{table}}

\section{Benchmark construction: difficulty tiers}
\label{sec:appendix}
Our difficulty tiers are not hand-set thresholds but a reproducible,
data-driven construction that we then \emph{independently} validate. Each
\{image, category\} pair is graded on three complementary axes
(Sec.~\ref{sec:difficulty}): scene occupancy (foreground ratio,
$\uparrow$~harder), target visibility (target-to-foreground ratio,
$\downarrow$~harder), and edit capacity (maximum addable count,
$\uparrow$~harder). Because the raw axes live on incomparable scales, we
rank-normalize each to a near-uniform $[0,1]$ score, sum them into a composite
difficulty in $[0,3]$, and split at its terciles into easy/medium/hard
(cut-points $1.27$ and $1.73$; observed scores span $0.08$--$2.94$). The
split is thus balanced by construction, with no arbitrary cut-offs.

\begin{figure}[!ht]
  \centering
  \includegraphics[width=0.85\linewidth]{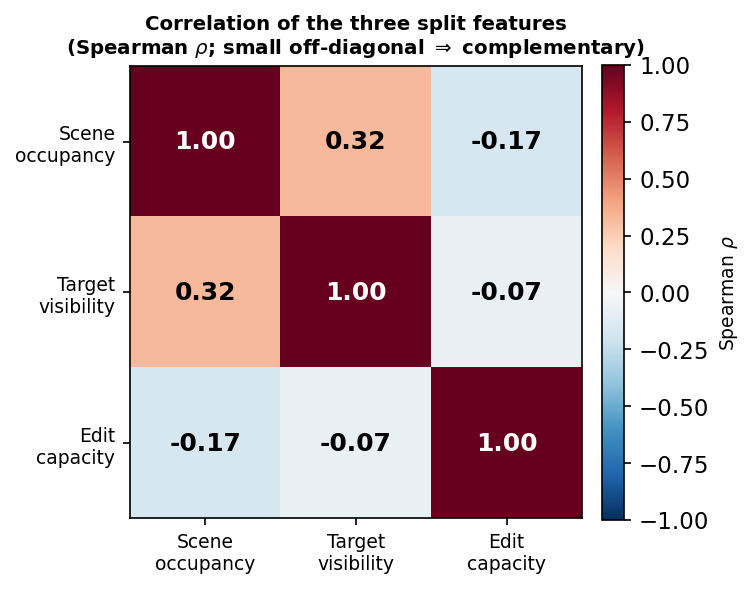}
  \caption{\textbf{Correlation of the three split features} (Spearman $\rho$).
  The scoring axes, scene occupancy (fg\_ratio), target visibility
  (target\_fg\_ratio), and edit capacity (max\_count), are only weakly
  inter-correlated ($|\rho|\!\le\!0.32$), so they capture complementary aspects
  of scene difficulty rather than restating one another.}
  \label{fig:diff-corr}
\end{figure}

The three split features are complementary rather than redundant: their
pairwise Spearman correlations are all small
(Fig.~\ref{fig:diff-corr}; $|\rho|\!\le\!0.32$), so each contributes independent
signal to the composite difficulty rather than restating another.

\begin{figure*}[t]
  \centering
  \includegraphics[width=\linewidth]{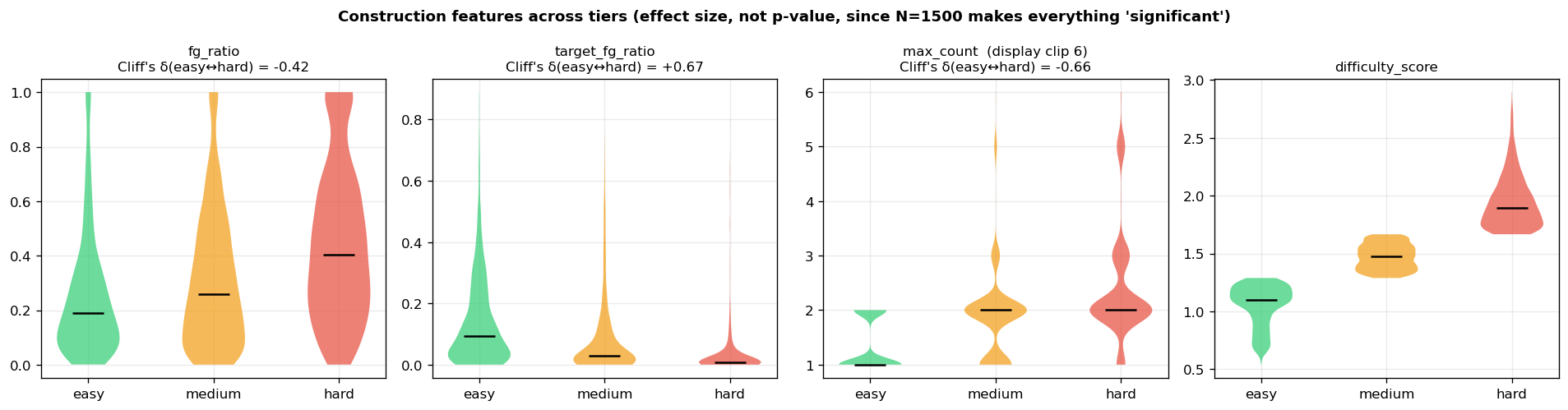}
  \caption{\textbf{The construction axes separate the tiers.} Per-tier
  distributions of the three axes and of the composite difficulty score, with
  Cliff's $\delta$ effect sizes (reported instead of $p$-values, which are
  uninformative at this sample size). Every axis shifts monotonically from easy
  to hard, and the difficulty score is strongly and cleanly ordered.}
  \label{fig:diff-violins}
\end{figure*}

Each axis is more than correlated with the label it helps define: all three
separate the tiers with a large effect size (Fig.~\ref{fig:diff-violins};
Cliff's $\delta$ between easy and hard of $-0.42$, $+0.67$, and $-0.66$), and
the composite score is cleanly ordered easy~$<$~medium~$<$~hard.

\begin{figure*}[t]
  \centering
  \includegraphics[width=\linewidth]{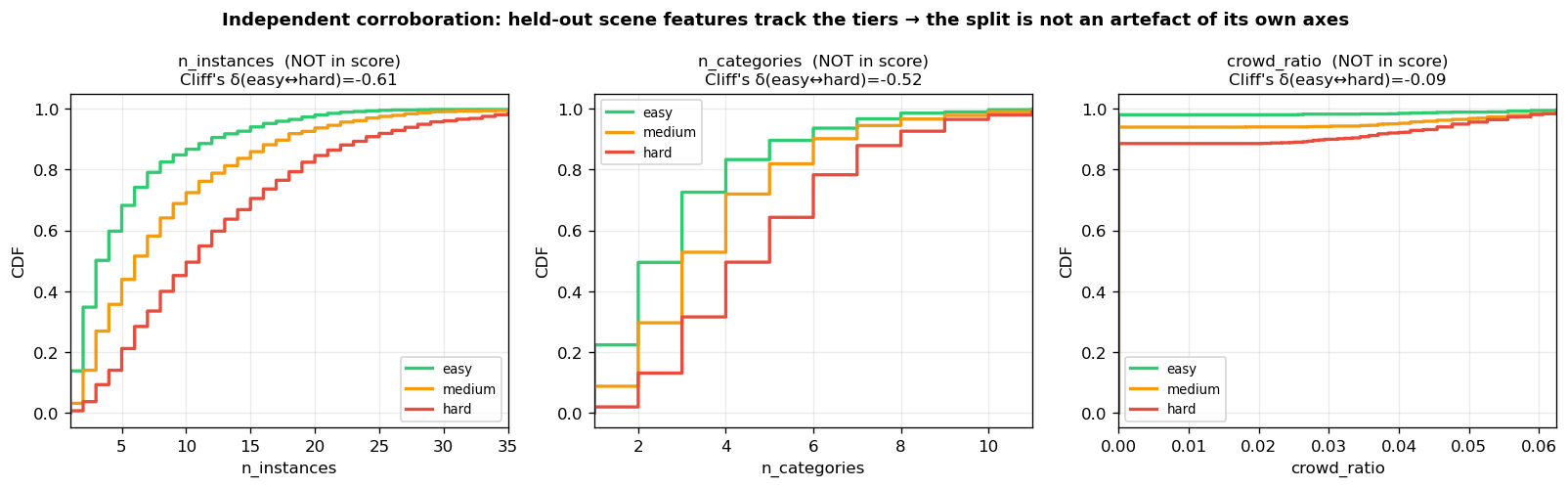}
  \caption{\textbf{The tiers are non-circular.} Per-tier CDFs of three held-out
  scene attributes that do \emph{not} enter the difficulty score. They still
  separate easy/medium/hard (Cliff's $\delta$ shown), so the split reflects real
  scene complexity, not the axes that define it. Crowd ratio ($\delta=-0.09$)
  separates least, as expected, since it is the most orthogonal to the
  construction.}
  \label{fig:diff-heldout}
\end{figure*}

Crucially, the split is \emph{not circular}. Three scene attributes that
\emph{never enter} the score, the number of instances, the number of
categories, and the crowd ratio, still track the tiers
(Fig.~\ref{fig:diff-heldout}; Cliff's $\delta=-0.61$, $-0.52$, $-0.09$): harder
tiers hold more instances and more categories, exactly as an \emph{independent}
notion of scene complexity would predict. The tiers thus capture genuine
difficulty rather than an artefact of their own axes.

\begin{figure}[!ht]
  \centering
  \includegraphics[width=0.82\linewidth]{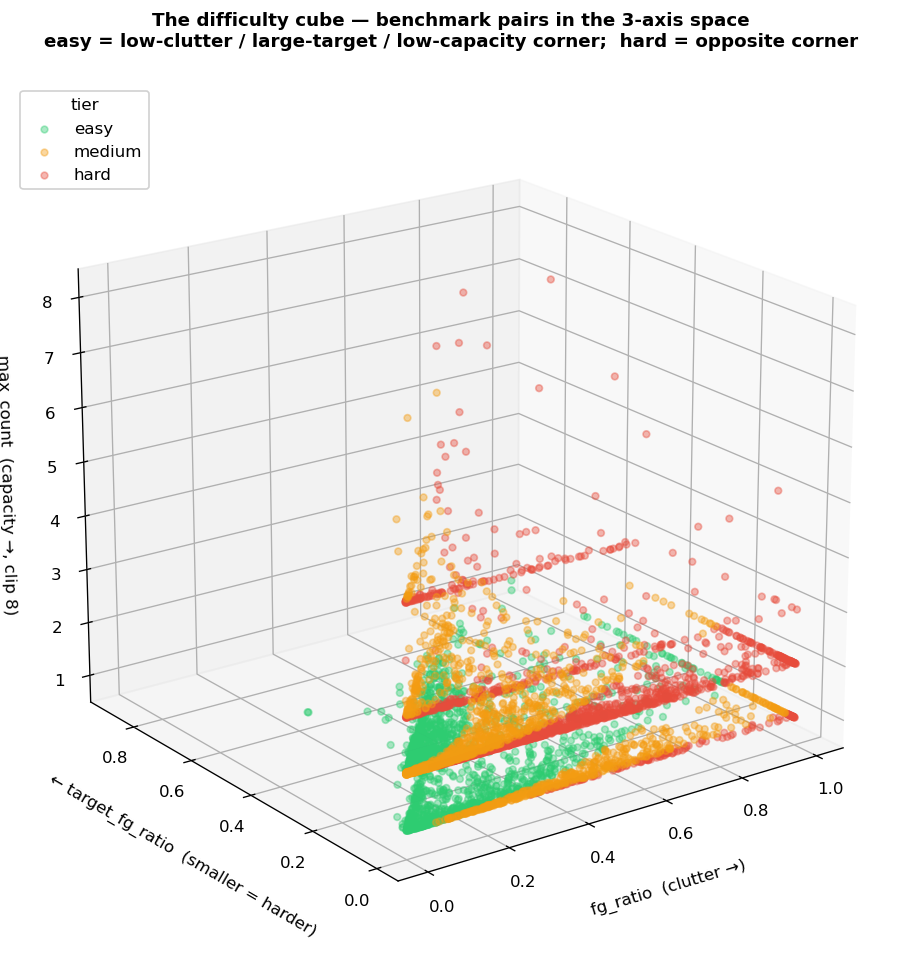}
  \caption{\textbf{The difficulty cube.} Benchmark pairs in the three-axis
  space, colored by tier. Easy pairs (green) cluster in the low-clutter,
  large-target, low-capacity corner; hard pairs (red) fill the opposite corner,
  with medium (orange) in between.}
  \label{fig:diff-cube}
\end{figure}

\begin{figure}[!ht]
  \centering
  \includegraphics[width=\linewidth]{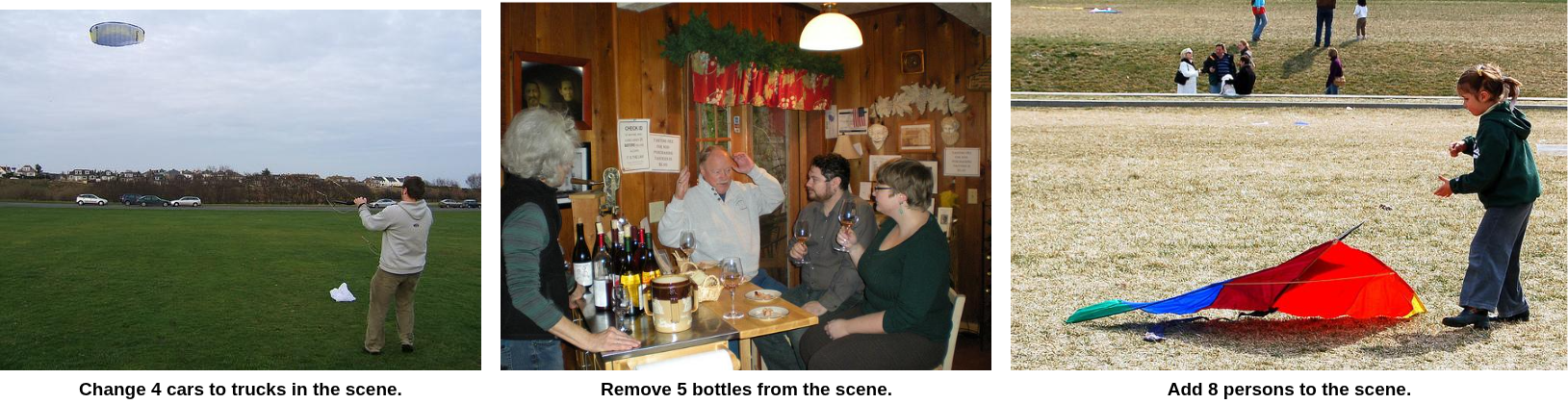}
  \caption{\textbf{Benchmark examples.} One \emph{add}, one \emph{remove},
  and one \emph{change} sample, each with the instruction shown beneath the
  image pair.}
  \label{fig:examples}
\end{figure}

\begin{figure}[!ht]
  \centering
  \includegraphics[width=\linewidth]{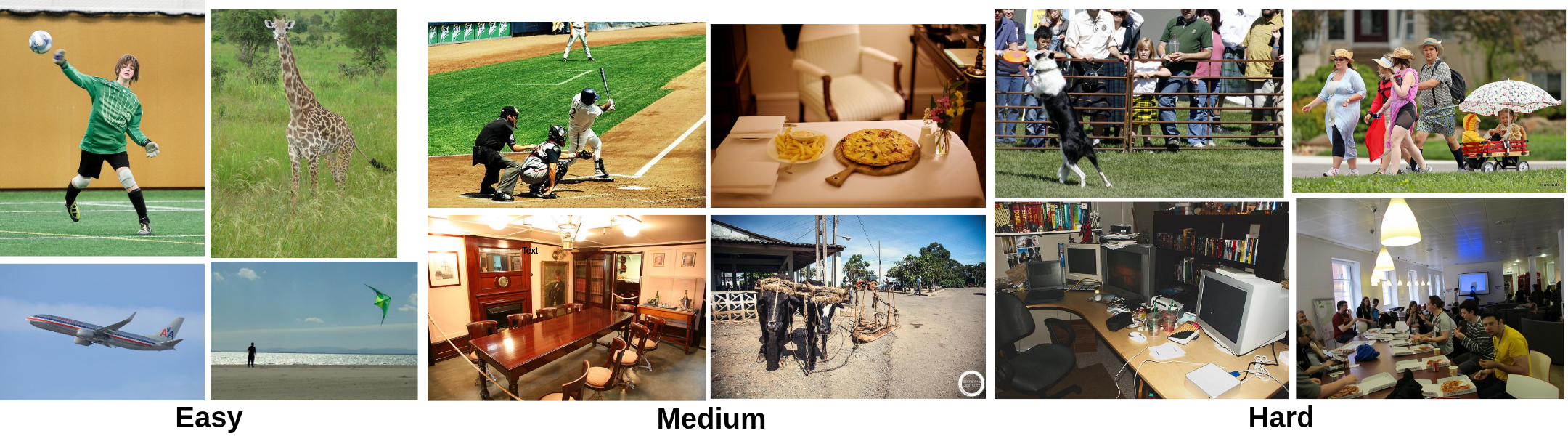}
  \caption{\textbf{Difficulty tiers.} Real examples for easy, medium, and
  hard. Difficulty is a tercile split of a composite score over three axes:
  foreground coverage, target count, and target-object scale. This makes
  precise what a ``complex scene'' is.}
  \label{fig:difficulty}
\end{figure}

Geometrically, the benchmark fills a difficulty ``cube'': the easy corner is
low-clutter, large-target, and low-capacity, and difficulty grows toward the
opposite corner (Fig.~\ref{fig:diff-cube}). One example of each operation
is shown in Fig.~\ref{fig:examples}, and the easy, medium, and hard
tiers in Fig.~\ref{fig:difficulty}.

\section{The evaluation protocol in action}
\label{sec:appendix-protocol}

\subsection{Notation}
\label{sec:notation}
Table~\ref{tab:notation} maps the pipeline outputs named in the figures to the
symbols used in the scoring equations of Sec.~\ref{sec:scoring}. Conservation
(Eqs.~\ref{eq:oc}, \ref{eq:rem}, \ref{eq:chg}) is normalized by the number of
original instances, so it measures the fraction of originals preserved: by
design, a single overwrite weighs more when few originals exist.

\begin{table}[!ht]
  \centering
  \footnotesize
  \setlength{\tabcolsep}{2.5pt}
  \renewcommand{\arraystretch}{0.95}
  \begin{tabular}{llll}
    \toprule
    Op. & Pipeline output & Symbol & Role in the Score \\
    \midrule
    Add    & $N^{\mathrm{added}}$     & $d$ & new instances; CF (Eq.~\ref{eq:cf}) \\
           & $N^{\mathrm{unmatched}}$ & $N_{\mathrm{rem}}$ & originals destroyed; OC (Eq.~\ref{eq:oc}) \\
    \midrule
    Remove & $N^{\mathrm{unmatched}}$ & $r$ & removed instances; CF (Eq.~\ref{eq:rem}) \\
           & $N^{\mathrm{added}}$     & $a$ & spurious additions; OC (Eq.~\ref{eq:rem}) \\
    \midrule
    Change & $N^{\mathrm{changed}}$   & $t$ & in-place changes; CF (Eq.~\ref{eq:chg}) \\
           & $N_{\mathrm{cls1}}-N_{\mathrm{cls1}}^{\mathrm{ip}}$ & $f_1$ & $\mathrm{cls1}$ deleted, not replaced; OC \\
           & $N_{\mathrm{cls2}}-N_{\mathrm{cls2}}^{\mathrm{ip}}$ & $f_2$ & $\mathrm{cls2}$ painted elsewhere; OC \\
    \midrule
    All    & instruction & $K$ & requested count; CF \\
           & ground truth & $N_{\mathrm{GT}}$ & original instances; OC denominator \\
    \bottomrule
  \end{tabular}
  \caption{\textbf{Notation.} Pipeline outputs named in the figures
  (Figs.~\ref{fig:add} and~\ref{fig:changepipe}) and the corresponding symbols
  in the scoring equations of Sec.~\ref{sec:scoring}. For change, the OC
  denominator is $N_{\mathrm{GT}}^{\mathrm{cls1}}$.}
  \label{tab:notation}
\end{table}

\subsection{Addition pipeline: per-sample walkthroughs}
\label{sec:addcase}
\begin{figure*}[t]
  \centering
  \includegraphics[width=\linewidth]{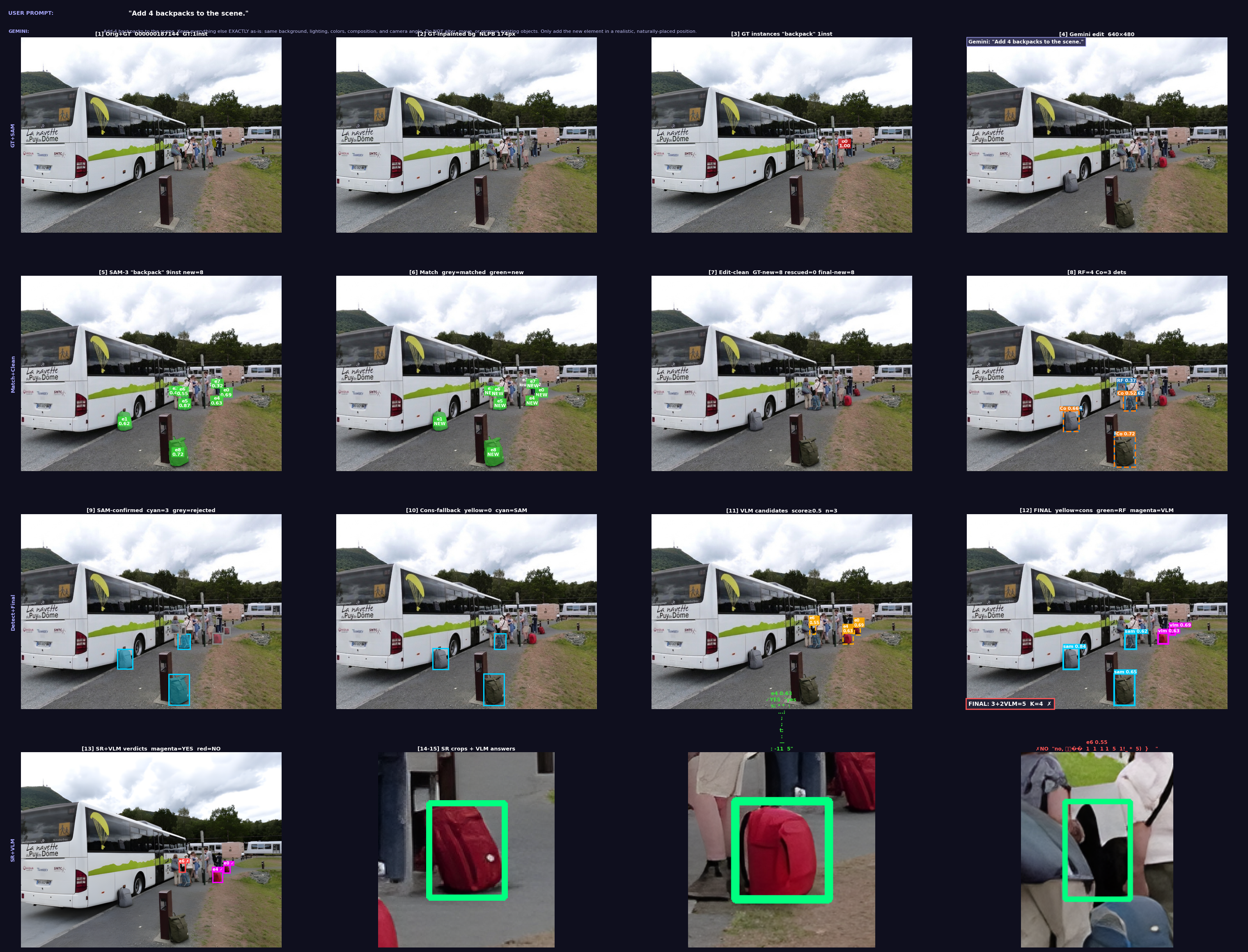}
  \caption{\textbf{Addition pipeline on a real edit (``add 4 backpacks'').}
  \emph{Row 1:} source$+$GT, NLPB-inpainted background, GT instances, and the
  Gemini edit. \emph{Row 2:} presence-gated SAM-3 segmentation, one-to-one
  matching (grey$=$matched, green$=$new), clean-image construction, and the two
  detectors' boxes. \emph{Row 3:} SAM-confirmed candidates, consensus fallback,
  VLM candidates, and the final verdict. \emph{Row 4:} the super-resolved crops
  sent to the constrained VLM verifier with its yes/no answers. The editor adds
  five backpacks for $K\!=\!4$; \name{} recovers the true count and penalizes
  it.}
  \label{fig:addcase}
\end{figure*}

\begin{figure*}[p]
  \centering
  \includegraphics[width=\linewidth]{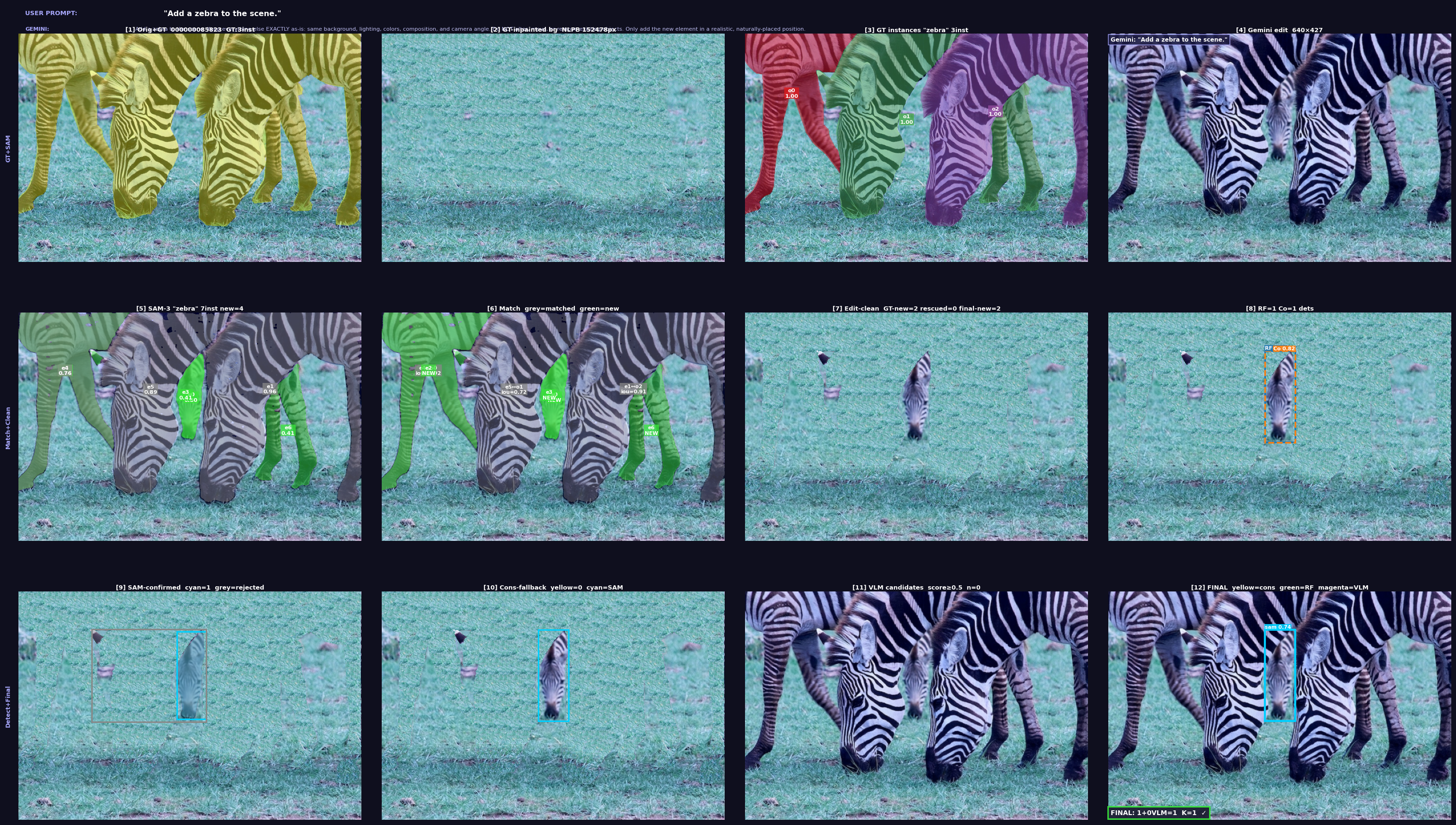}\\[4pt]
  \includegraphics[width=\linewidth]{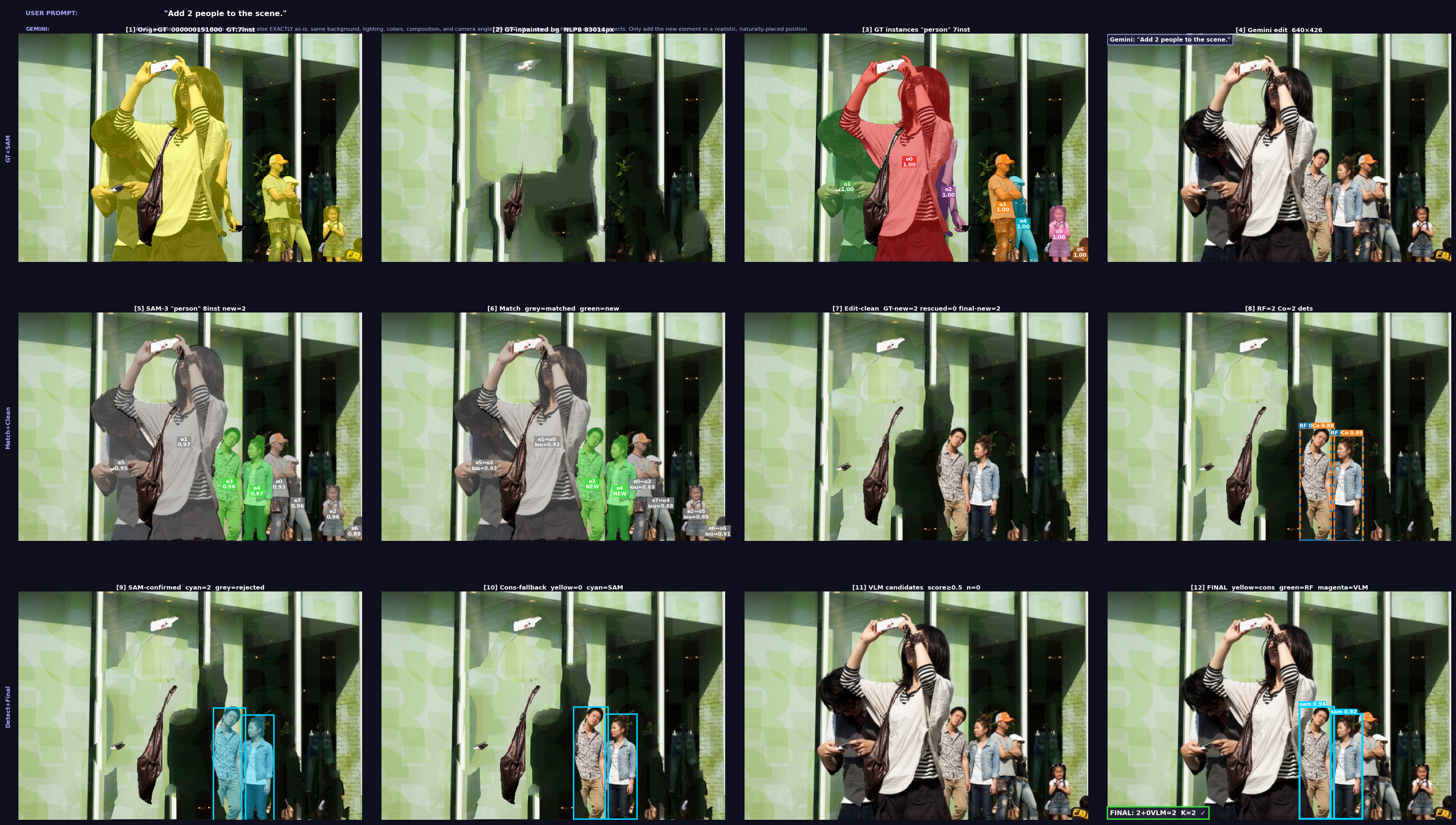}
  \caption{\textbf{Correct additions (I).} ``Add a zebra'' (top, one added for
  $K\!=\!1$) and ``add 2 people'' (bottom, two for $K\!=\!2$). The added zebra
  sits \emph{behind} the grazing animals, yet it is still segmented, isolated by
  inpainting, and confirmed by both detectors; \name{} credits the edit.}
  \label{fig:addgallery1}
\end{figure*}

\begin{figure*}[p]
  \centering
  \includegraphics[width=\linewidth]{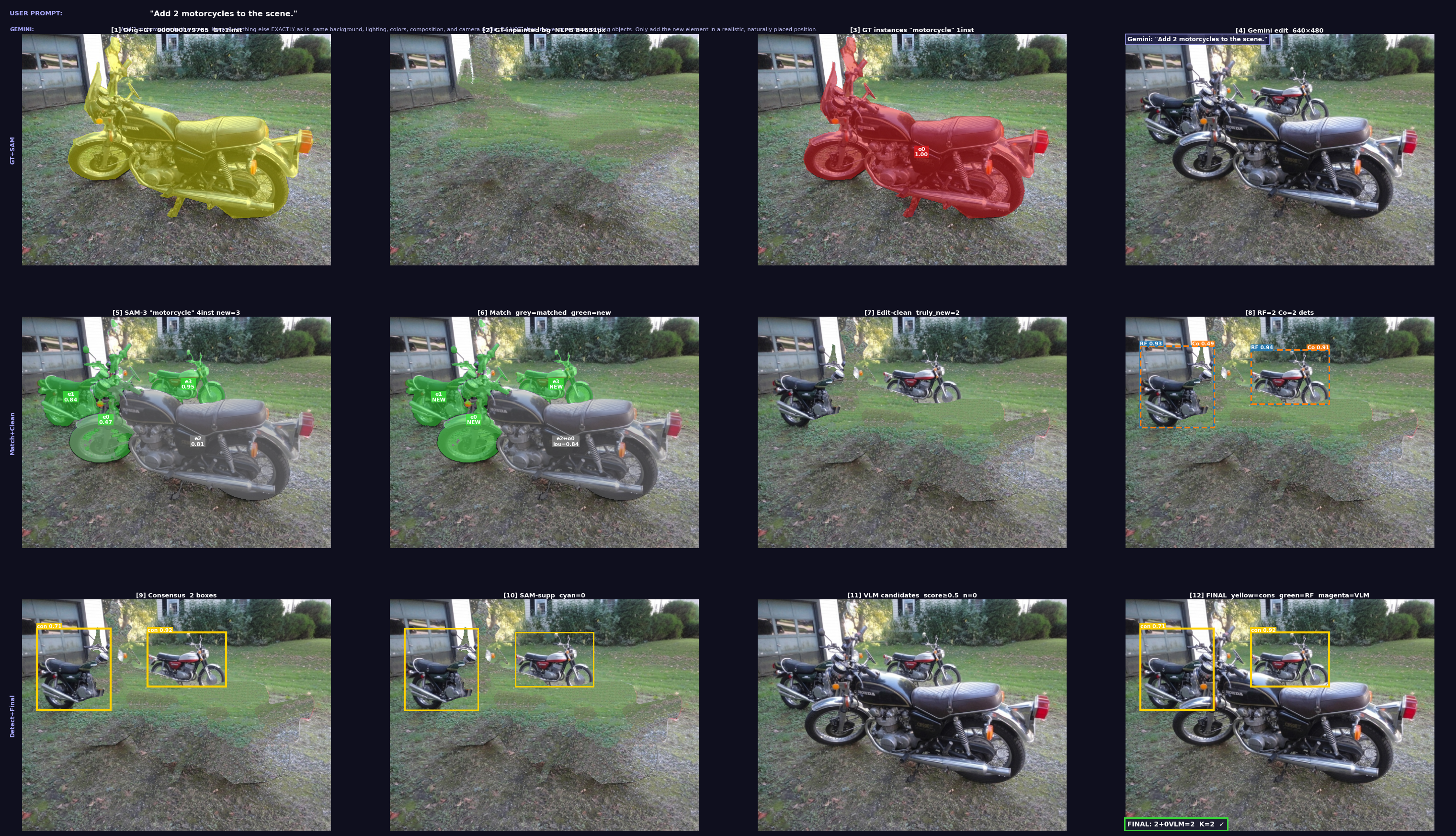}\\[4pt]
  \includegraphics[width=\linewidth]{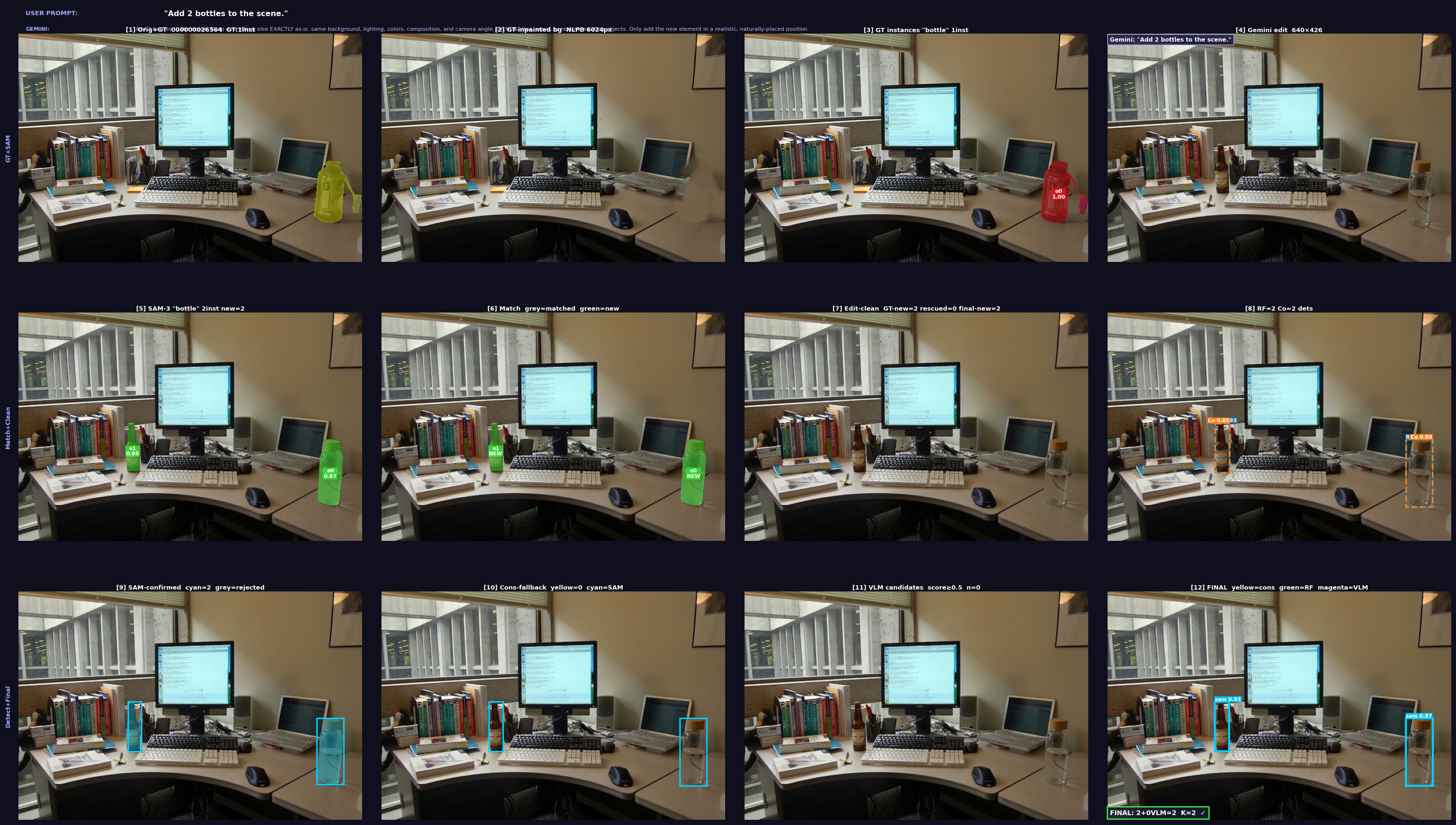}
  \caption{\textbf{Correct additions (II).} ``Add 2 motorcycles'' (top) and
  ``add 2 bottles'' (bottom), both correctly counted at $K\!=\!2$. The new
  motorcycles sit \emph{behind} the foreground one and are still recovered,
  while the bottles are added among the existing desk clutter.}
  \label{fig:addgallery2}
\end{figure*}

\begin{figure*}[p]
  \centering
  \includegraphics[width=\linewidth]{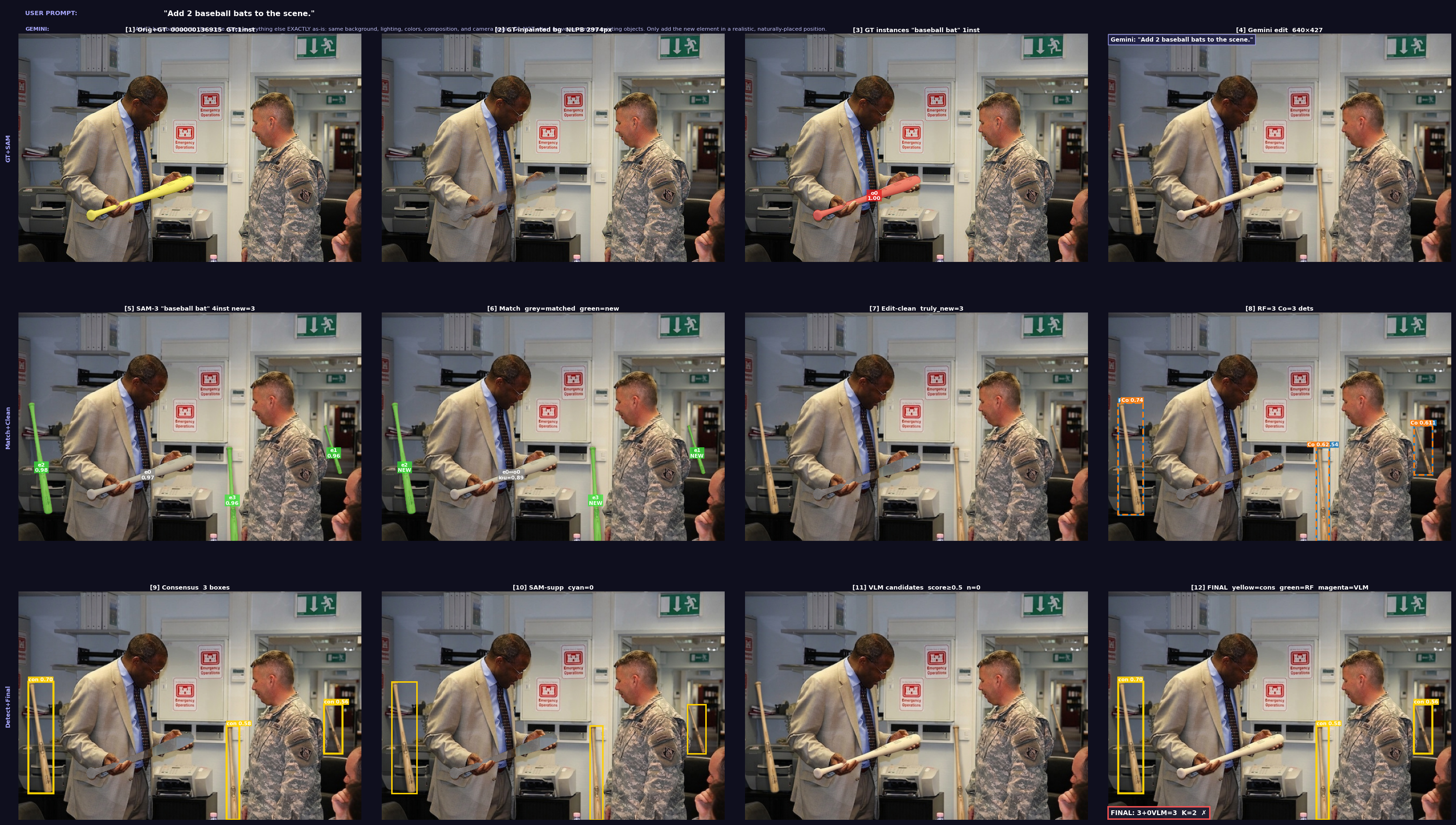}\\[4pt]
  \includegraphics[width=\linewidth]{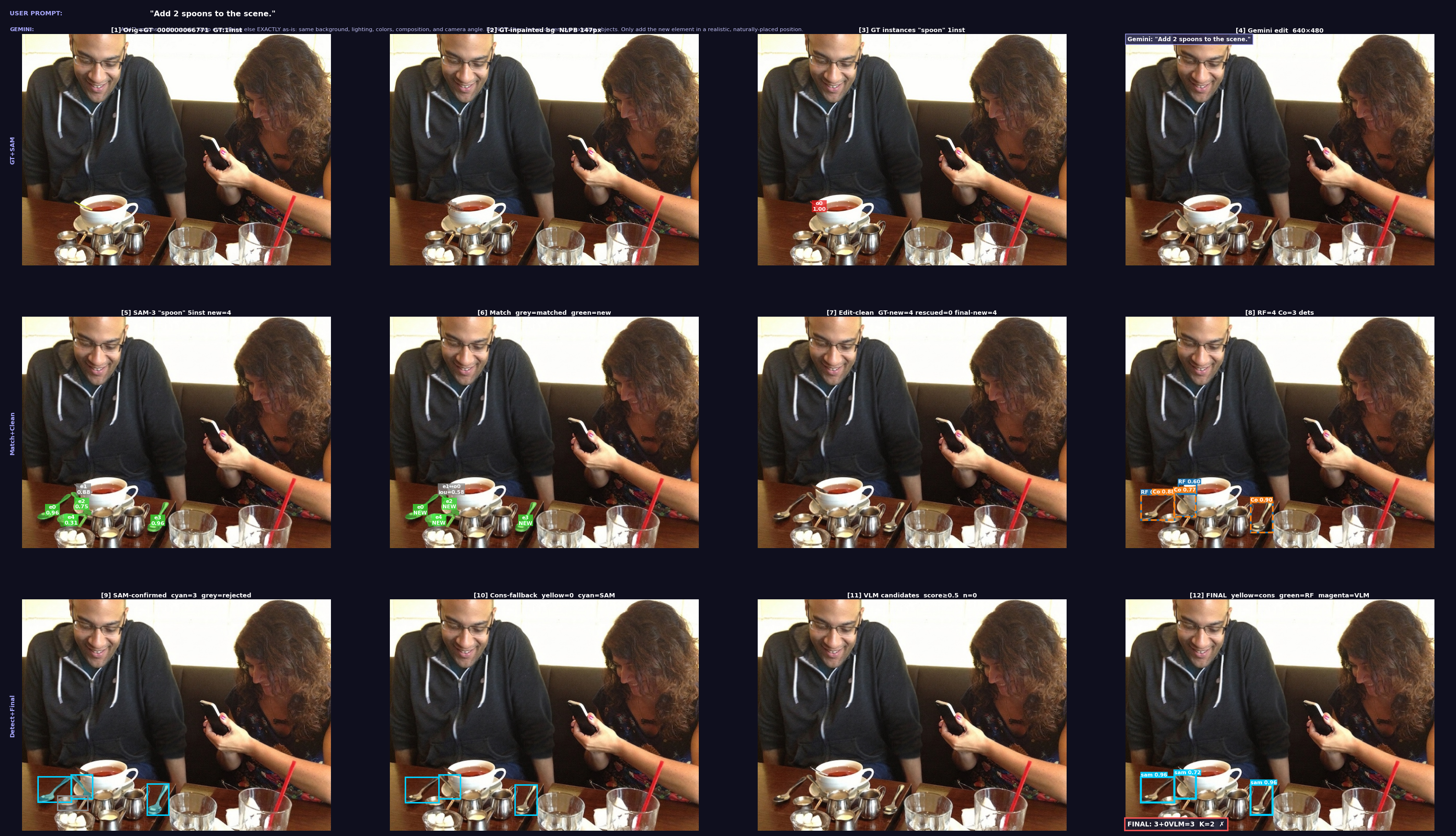}
  \caption{\textbf{Caught over-additions.} ``Add 2 baseball bats'' (top) and
  ``add 2 spoons'' (bottom) each yield three instances; \name{} counts the
  surplus and the count-fidelity term penalizes the miscount, which a holistic
  or VLM-judge score would overlook.}
  \label{fig:addgallery3}
\end{figure*}

Figures~\ref{fig:addcase}--\ref{fig:addhard} trace the addition pipeline
(Sec.~\ref{sec:add}) panel by panel on real Gemini edits, making the
otherwise-abstract stages concrete and showing the protocol across categories
and target counts. Figure~\ref{fig:addcase} is the most complete: an
``add 4 backpacks'' edit that exercises every stage, including the rare
ambiguity-resolution step where unconfirmed candidates are super-resolved and
passed to the constrained VLM verifier (bottom row). The editor over-adds, so
\name{} counts five backpacks for $K\!=\!4$ and penalizes the miscount, exactly
the failure a count-blind metric would miss.

The same pipeline runs unchanged across diverse categories and counts
(Fig.~\ref{fig:addgallery1}--\ref{fig:addgallery3}). It credits edits that
reach the requested count and catches those that do not, every verdict grounded
in the same segmentation, matching, and dual-detector confirmation.

The protocol holds at high target counts, the regime where editors fail most
(Fig.~\ref{fig:addhard}).

\subsection{Repairing incomplete ground truth}
\label{sec:gtcompletion}
\begin{figure*}[p]
  \centering
  \includegraphics[width=0.9\linewidth]{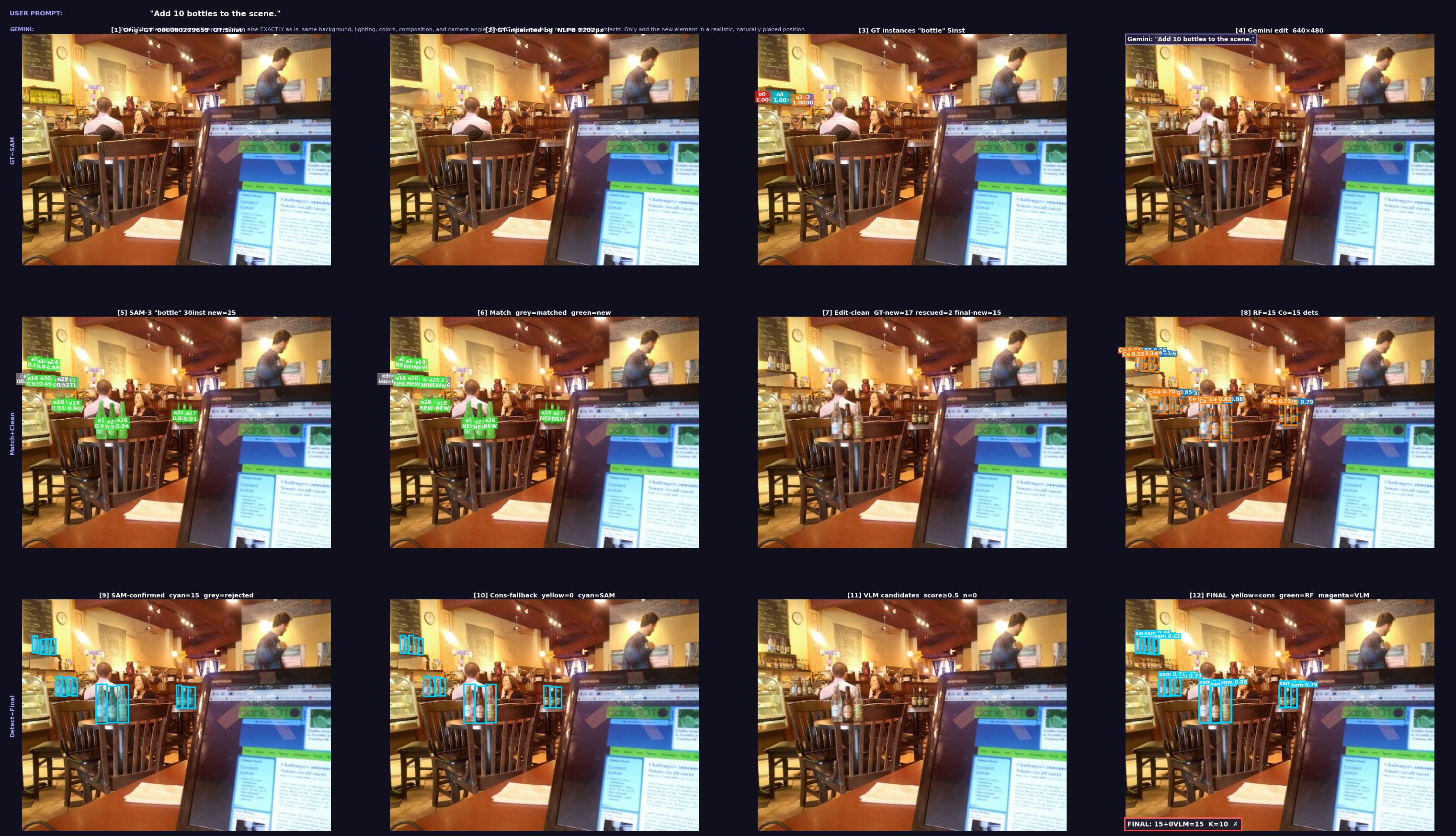}
  \caption{\textbf{High-count addition (``add 10 bottles'').} A stress case at
  the top of the count range. The editor packs in fifteen bottles;
  presence-gated segmentation and dual-detector confirmation resolve the dense,
  overlapping instances, and \name{} reports $15$ against $K\!=\!10$, a miscount
  a holistic or VLM-judge metric would overlook.}
  \label{fig:addhard}

  \vspace{12pt}
  \centering
  \includegraphics[width=0.9\linewidth]{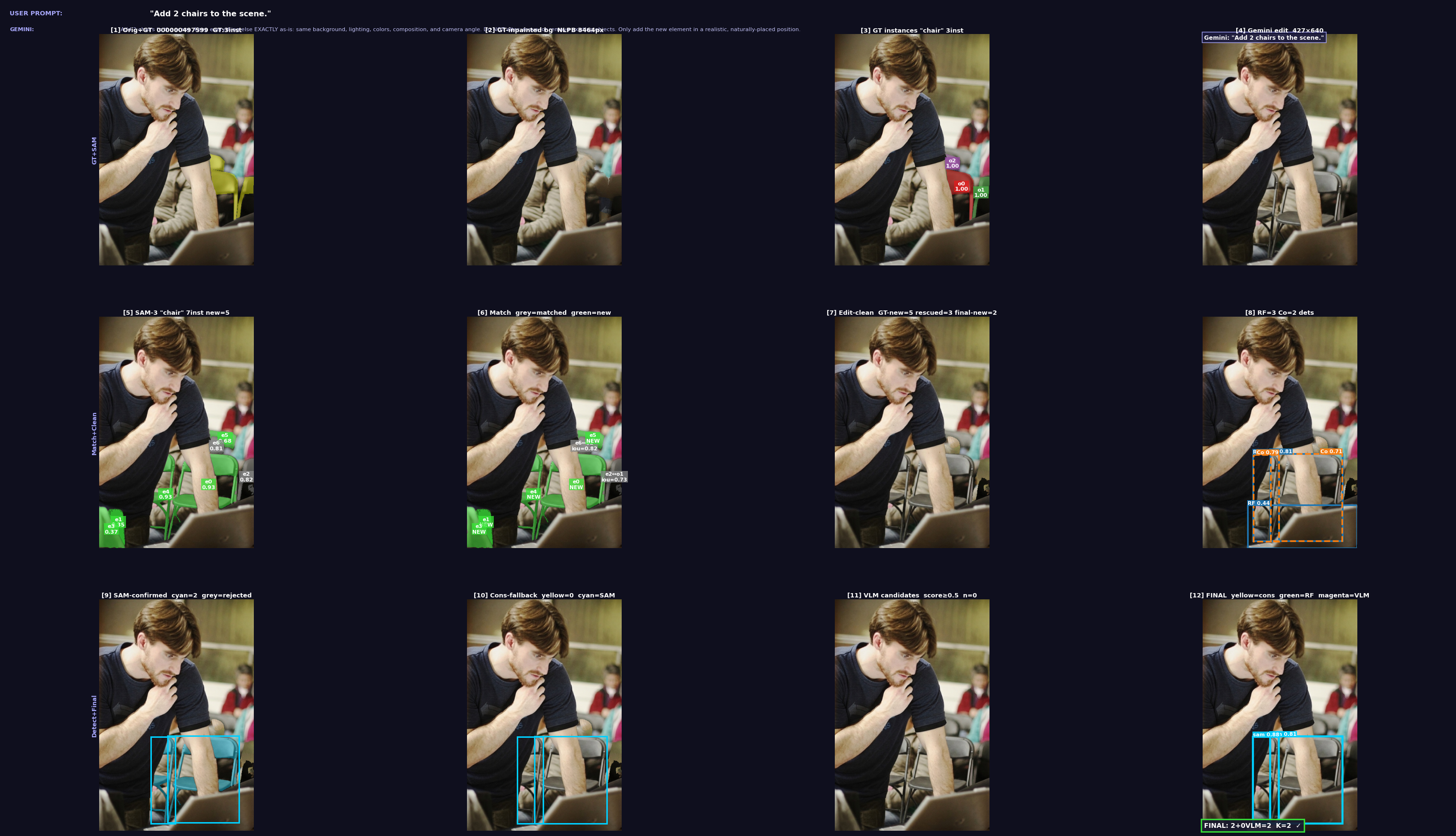}
  \caption{\textbf{Ground-truth completion on ``add 2 chairs''.} The background
  chairs already exist in the source image but are absent from the panoptic
  annotation, so segmentation of the edit initially flags them as new. Because
  \name{} also segments the \emph{source} image, the cross-check finds their
  counterparts there and reassigns them to \emph{existing} (rescued), leaving
  only the genuinely added chairs in the count. Without this step, missing
  annotations would be mistaken for additions and inflate the count.}
  \label{fig:gtcompletion}
\end{figure*}

The panoptic annotations are strong but not perfect: some instances are simply
not labelled. A naive pipeline would then count a real, pre-existing object as
\emph{newly added}, inflating the realized count. \name{} guards against this by
segmenting the \emph{source} image as well, not just the edit: every candidate
flagged as new is cross-checked against the source segmentation, and if a
confident counterpart is present in the original it is reassigned to
\emph{existing} (``rescued'') rather than counted as new
(Sec.~\ref{sec:add}). Figure~\ref{fig:gtcompletion} shows a clear case on an
``add 2 chairs'' edit: the chairs in the background were present in the original
but missing from the annotation, so SAM-3 first marks them as new; the
source-image cross-check finds their counterparts and rescues them, so only the
genuinely added chairs are counted.

\subsection{Matching under generative drift}
\label{sec:matchcase}
Figure~\ref{fig:matchcar} shows why exact-count editing needs more than a direct
judge. The edit inserts two cars so naturally that the before/after pair looks
like an ordinary street: a VLM judge sees a plausible result and cannot say
what, or how much, changed. \name{}'s geometry-and-appearance matcher
(Sec.~\ref{sec:add}) instead places every edited instance against the source.
It confirms the pre-existing car as \emph{matched} (e1) and flags the two
inserted cars as genuinely \emph{new} (e0 and e2, with high segmentation
scores), recovering the exact added count that a direct judge would miss.

\subsection{The change operation}
\label{sec:changecase}
\begin{figure*}[p]
  \centering
  \includegraphics[width=0.48\linewidth]{figures/benchmark/example_match_car_before}\hfill
  \includegraphics[width=0.48\linewidth]{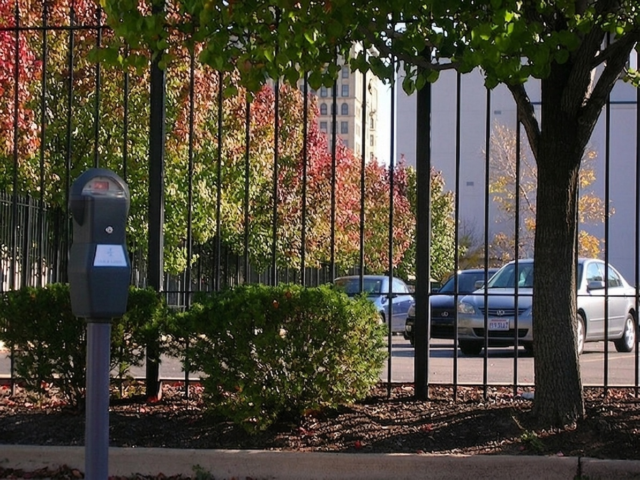}\\[5pt]
  \includegraphics[width=0.72\linewidth]{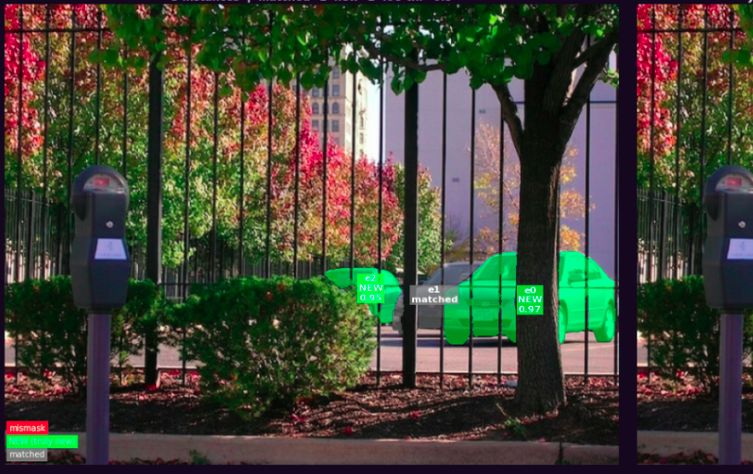}
  \caption{\textbf{Matching catches changes a judge cannot see.} \emph{Top:}
  source (left) and edited (right) images for ``add 2 cars''; the two added cars
  blend in so well that the before/after difference is easy to miss, and a
  direct VLM judge cannot tell what changed. \emph{Bottom:} \name{}'s
  geometry-and-appearance matcher resolves it exactly, marking the pre-existing
  car \emph{matched} (e1) and the two inserted cars \emph{new} (e0, e2). The
  one-to-one matching is what turns a near-imperceptible edit into a precise,
  grounded count.}
  \label{fig:matchcar}

  \vspace{12pt}
  \centering
  \includegraphics[width=0.5\linewidth]{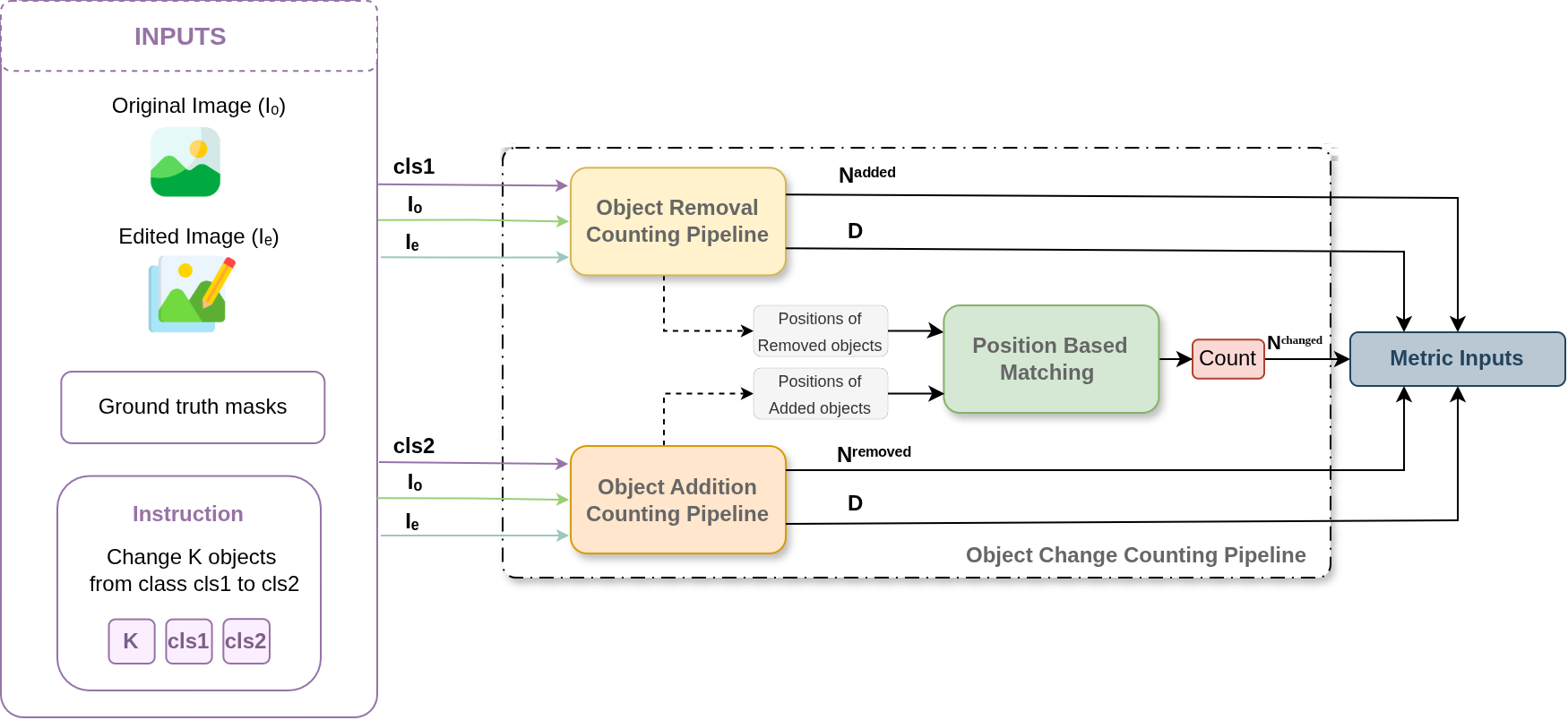}
  \caption{\textbf{`Change' pipeline.} The remove pipeline runs on
  $\mathrm{cls1}$ and the add pipeline on $\mathrm{cls2}$; matching the
  removed and added positions yields the matched changes
  $N^{\mathrm{changed}}$.}
  \label{fig:changepipe}

\end{figure*}
\begin{figure*}[t]
  \centering
  \includegraphics[width=\linewidth]{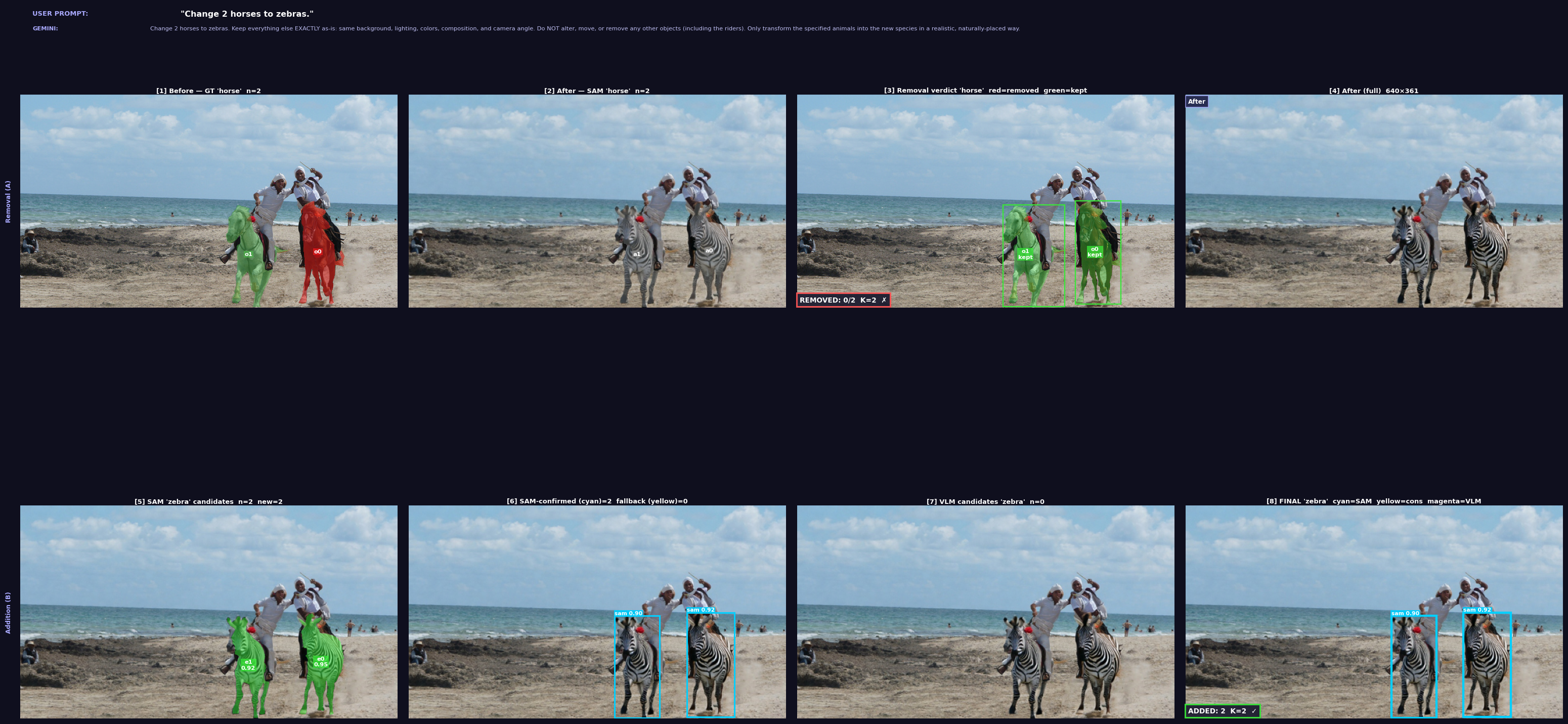}
  \caption{\textbf{Change pipeline on a real edit (``change 2 horses to
  zebras'').} Each instance is colored by its verdict as the pipeline proceeds:
  the source horses are segmented and matched, the edited zebras are detected
  and confirmed by the two detectors, and the removed-source and added-target
  positions are matched so each change is credited only where a horse originally
  stood. This position-aware coupling is what prevents a delete-here-paint-there
  edit from passing as a change.}
  \label{fig:change}
\end{figure*}

\afterpage{\clearpage}
Figure~\ref{fig:changepipe} summarizes the change pipeline. Figure~\ref{fig:change} walks the change pipeline (Sec.~\ref{sec:change})
through a real ``change 2 horses to zebras'' edit. The remove branch tracks the
source $\mathrm{cls1}$ (horses) and the add branch the new $\mathrm{cls2}$
(zebras); matching their positions credits a transformation only where it
occurred \emph{in place}, where a horse was, which is what separates a genuine
change from a delete-here-paint-there edit.

\section{Additional experimental analysis}
\label{sec:appendix-experiments}

\subsection{Leaderboard computation}
\label{sec:appendix-leaderboard}
Every number in Table~\ref{tab:leaderboard} is obtained from the per-edit
verdicts in four steps, using the cell sizes $N_c$ of
Table~\ref{tab:composition} ($\sum_c N_c=10{,}944$).
\begin{enumerate}\itemsep0pt
\item \textbf{Cell factors.} For each editor and each of the nine
operation$\times$tier cells $c$, $\mathrm{CF}_c$, $\mathrm{SC}_c$ and
$\mathrm{OC}_c$ are the means of the per-edit factors
(Sec.~\ref{sec:scoring}) over the $N_c$ edits of the cell
(Table~\ref{tab:factors}).
\item \textbf{Cell Score.} $\mathrm{Score}_c=\mathrm{CF}_c\cdot\mathrm{SC}_c\cdot\mathrm{OC}_c$,
computed at full precision and rounded half-up to two decimals for the E/M/H
columns of Table~\ref{tab:leaderboard}.
\item \textbf{Overall factors.} $\mathrm{CF}=\sum_c N_c\,\mathrm{CF}_c/\sum_c N_c$,
\ie the mean of the per-edit CF over all $10{,}944$ edits, and likewise for
SC and OC.
\item \textbf{Overall Score.} $\mathrm{Score}=\mathrm{CF}\cdot\mathrm{SC}\cdot\mathrm{OC}$,
the product of the three overall averages. It is \emph{not} the mean of the
nine cell Scores, which is why it can sit below many of them.
\end{enumerate}
We aggregate this way, rather than averaging the per-edit Scores of
Eq.~\ref{eq:score}, so that every reported Score is exactly the product of the
CF, SC, and OC columns printed next to it: a reader can trace any difference
between editors to the factor that causes it. The per-edit Score of
Eq.~\ref{eq:score} remains the quantity each individual edit receives.
Because the overall factors are computed from the per-edit values, recomputing
them from the three-decimal cells of Table~\ref{tab:factors} can differ in the
fourth decimal.

\noindent\textbf{Worked example (Seedream~4.0).} For the Add/Easy cell,
$\mathrm{Score}_c=0.762\times0.922\times0.782=0.549\to0.55$; for
Change/Hard, $0.469\times0.607\times0.476=0.136\to0.14$.
Weighting the cell factors by $N_c$ gives
$\mathrm{CF}=6605.6/10{,}944=0.6036$,
$\mathrm{SC}=8689.0/10{,}944=0.7940$ and
$\mathrm{OC}=7174.6/10{,}944=0.6556$, hence
$\mathrm{Score}=0.6036\times0.7940\times0.6556=0.3142\to0.31$.

\noindent\textbf{Exact-count success rate.} As a complement to the Score, the
share of edits whose count is exactly right ($\mathrm{CF}=1$) is $0.43$ for
Seedream~4.0, $0.41$ for Gemini~2.5, $0.38$ for GPT-Image-1, $0.21$ for
Qwen-Image-Edit, $0.19$ for FLUX.1 Kontext, $0.13$ for Step1X-Edit, and $0.09$
for BAGEL: even the best editor delivers the requested number on fewer than
half of the edits, with the same ordering as the Score.

\begin{table*}[!t]
  \centering
  \footnotesize
  \setlength{\tabcolsep}{4.5pt}
  \renewcommand{\arraystretch}{0.95}
  \begin{tabular}{llccccccccc}
    \toprule
    & & \multicolumn{3}{c}{Add} & \multicolumn{3}{c}{Remove} & \multicolumn{3}{c}{Change} \\
    \cmidrule(lr){3-5}\cmidrule(lr){6-8}\cmidrule(lr){9-11}
    Model & Factor & E & M & H & E & M & H & E & M & H \\
    \midrule
    & $N_c$ & 1{,}159 & 1{,}363 & 1{,}374 & 1{,}146 & 1{,}340 & 1{,}308 & 1{,}000 & 1{,}128 & 1{,}126 \\
    \midrule
    \multirow{4}{*}{Seedream 4.0~\cite{seedream4}}
      & CF & 0.762 & 0.662 & 0.579 & 0.684 & 0.563 & 0.513 & 0.677 & 0.541 & 0.469 \\
      & SC & 0.922 & 0.852 & 0.776 & 0.901 & 0.800 & 0.724 & 0.839 & 0.726 & 0.607 \\
      & OC & 0.782 & 0.745 & 0.633 & 0.745 & 0.686 & 0.530 & 0.677 & 0.624 & 0.476 \\
      & $\mathrm{CF}\!\cdot\!\mathrm{SC}\!\cdot\!\mathrm{OC}$ & 0.5494 & 0.4202 & 0.2844 & 0.4591 & 0.3090 & 0.1968 & 0.3845 & 0.2451 & 0.1355 \\
    \midrule
    \multirow{4}{*}{Gemini 2.5 Flash Image~\cite{geminiflash}}
      & CF & 0.710 & 0.643 & 0.554 & 0.676 & 0.591 & 0.493 & 0.628 & 0.567 & 0.465 \\
      & SC & 0.885 & 0.843 & 0.768 & 0.863 & 0.806 & 0.728 & 0.756 & 0.684 & 0.579 \\
      & OC & 0.823 & 0.717 & 0.620 & 0.764 & 0.634 & 0.513 & 0.663 & 0.570 & 0.450 \\
      & $\mathrm{CF}\!\cdot\!\mathrm{SC}\!\cdot\!\mathrm{OC}$ & 0.5171 & 0.3886 & 0.2638 & 0.4457 & 0.3020 & 0.1841 & 0.3148 & 0.2211 & 0.1212 \\
    \midrule
    \multirow{4}{*}{GPT-Image-1~\cite{gptimage1}}
      & CF & 0.742 & 0.633 & 0.528 & 0.660 & 0.530 & 0.469 & 0.618 & 0.474 & 0.414 \\
      & SC & 0.843 & 0.814 & 0.744 & 0.809 & 0.773 & 0.688 & 0.757 & 0.707 & 0.619 \\
      & OC & 0.817 & 0.708 & 0.629 & 0.710 & 0.592 & 0.528 & 0.652 & 0.538 & 0.429 \\
      & $\mathrm{CF}\!\cdot\!\mathrm{SC}\!\cdot\!\mathrm{OC}$ & 0.5110 & 0.3648 & 0.2471 & 0.3791 & 0.2425 & 0.1704 & 0.3050 & 0.1803 & 0.1099 \\
    \midrule
    \multirow{4}{*}{Qwen-Image-Edit~\cite{qwenimage}}
      & CF & 0.566 & 0.476 & 0.368 & 0.472 & 0.443 & 0.333 & 0.432 & 0.371 & 0.303 \\
      & SC & 0.873 & 0.787 & 0.675 & 0.807 & 0.696 & 0.560 & 0.742 & 0.626 & 0.474 \\
      & OC & 0.704 & 0.611 & 0.511 & 0.672 & 0.570 & 0.452 & 0.567 & 0.481 & 0.359 \\
      & $\mathrm{CF}\!\cdot\!\mathrm{SC}\!\cdot\!\mathrm{OC}$ & 0.3479 & 0.2289 & 0.1269 & 0.2560 & 0.1757 & 0.0843 & 0.1817 & 0.1117 & 0.0516 \\
    \midrule
    \multirow{4}{*}{FLUX.1 Kontext~\cite{fluxkontext}}
      & CF & 0.511 & 0.454 & 0.349 & 0.485 & 0.395 & 0.318 & 0.379 & 0.327 & 0.288 \\
      & SC & 0.828 & 0.713 & 0.622 & 0.807 & 0.682 & 0.585 & 0.720 & 0.581 & 0.472 \\
      & OC & 0.707 & 0.612 & 0.548 & 0.616 & 0.512 & 0.445 & 0.518 & 0.451 & 0.354 \\
      & $\mathrm{CF}\!\cdot\!\mathrm{SC}\!\cdot\!\mathrm{OC}$ & 0.2991 & 0.1981 & 0.1190 & 0.2411 & 0.1379 & 0.0828 & 0.1414 & 0.0857 & 0.0481 \\
    \midrule
    \multirow{4}{*}{Step1X-Edit~\cite{step1xedit}}
      & CF & 0.428 & 0.367 & 0.353 & 0.373 & 0.345 & 0.311 & 0.315 & 0.290 & 0.257 \\
      & SC & 0.806 & 0.736 & 0.639 & 0.743 & 0.646 & 0.555 & 0.647 & 0.539 & 0.440 \\
      & OC & 0.682 & 0.585 & 0.440 & 0.635 & 0.522 & 0.394 & 0.535 & 0.402 & 0.290 \\
      & $\mathrm{CF}\!\cdot\!\mathrm{SC}\!\cdot\!\mathrm{OC}$ & 0.2353 & 0.1580 & 0.0992 & 0.1760 & 0.1163 & 0.0680 & 0.1090 & 0.0628 & 0.0328 \\
    \midrule
    \multirow{4}{*}{BAGEL~\cite{bagel}}
      & CF & 0.382 & 0.331 & 0.306 & 0.330 & 0.280 & 0.270 & 0.304 & 0.260 & 0.240 \\
      & SC & 0.800 & 0.713 & 0.575 & 0.763 & 0.643 & 0.490 & 0.675 & 0.549 & 0.388 \\
      & OC & 0.642 & 0.534 & 0.453 & 0.571 & 0.461 & 0.389 & 0.507 & 0.390 & 0.321 \\
      & $\mathrm{CF}\!\cdot\!\mathrm{SC}\!\cdot\!\mathrm{OC}$ & 0.1962 & 0.1260 & 0.0797 & 0.1438 & 0.0830 & 0.0515 & 0.1040 & 0.0557 & 0.0299 \\
    \bottomrule
  \end{tabular}
  \caption{\textbf{Per-cell factors behind Table~\ref{tab:leaderboard}.} Mean
  count fidelity (CF), scene consistency (SC), and original-category
  conservation (OC) per editor, operation, and difficulty tier, with the cell
  size $N_c$ and the cell Score $\mathrm{CF}_c\cdot\mathrm{SC}_c\cdot\mathrm{OC}_c$; rounding the exact product
  half-up to two decimals gives the E/M/H columns of
  Table~\ref{tab:leaderboard}.}
  \label{tab:factors}
\end{table*}

\subsection{Protocol ablation}
\label{sec:appendix-ablation}
Table~\ref{tab:ablation} gives the full component ablation analysed in
Sec.~\ref{sec:ablations}: each component of the evaluation protocol is removed
one at a time and exact-count agreement against the human count is re-measured
on the 600-image set. Every component is load-bearing, and the matcher's
one-to-one alignment is by far the most important (agreement collapses from
$0.82$ to $0.36$ without it), while the constrained VLM verifier is the least
important ($0.82$ to $0.79$), confirming that accuracy rests on grounded
geometry and detection rather than on the VLM.

\begin{table}[t]
  \centering
  \footnotesize
  \setlength{\tabcolsep}{5pt}
  \renewcommand{\arraystretch}{0.95}
  \begin{tabular}{lc}
    \toprule
    Variant & Acc.\ vs.\ human \\
    \midrule
    \textbf{Full \name{}}                        & \textbf{0.82} \\
    w/o ambiguity block (VLM verifier)           & 0.79 \\
    single detector (avg.\ RF-only, Co-only)     & 0.71 \\
    w/o inpainting (clean-image isolation)       & 0.70 \\
    w/o ground-truth completion                  & 0.69 \\
    w/o one-to-one alignment                     & 0.36 \\
    \bottomrule
  \end{tabular}
  \caption{\textbf{Ablation of the evaluation protocol.} Exact-count accuracy
  against the human count, removing one component at a time (600-image
  set; analysis in Sec.~\ref{sec:ablations}).}
  \label{tab:ablation}
\end{table}

\subsection{Human-evaluation set composition}
\label{sec:hecomp}
\begin{figure*}[t]
  \centering
  \includegraphics[width=\linewidth]{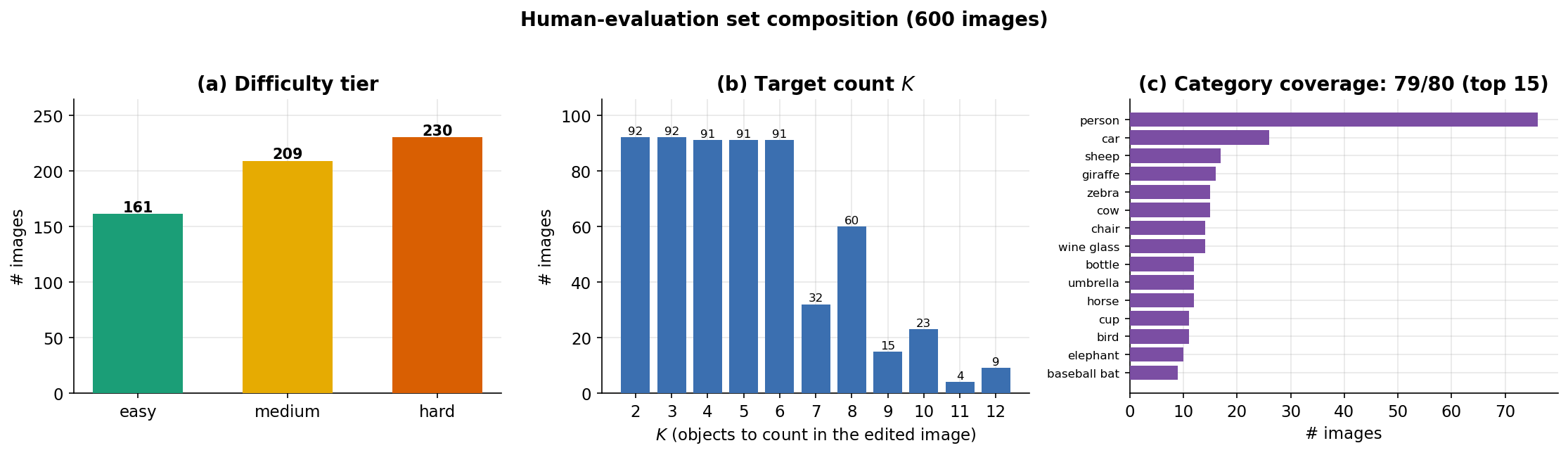}
  \caption{\textbf{Composition of the 600-image human-evaluation set.}
  (a) Difficulty-tier distribution (easy/medium/hard). (b) Target-count $K$
  distribution, concentrated at the low-to-mid counts and thinning at the
  hardest high counts. (c) Category coverage: 79 of 80 COCO thing-categories.}
  \label{fig:hecomp}
\end{figure*}

\begin{figure}[!ht]
  \centering
  \includegraphics[width=\linewidth]{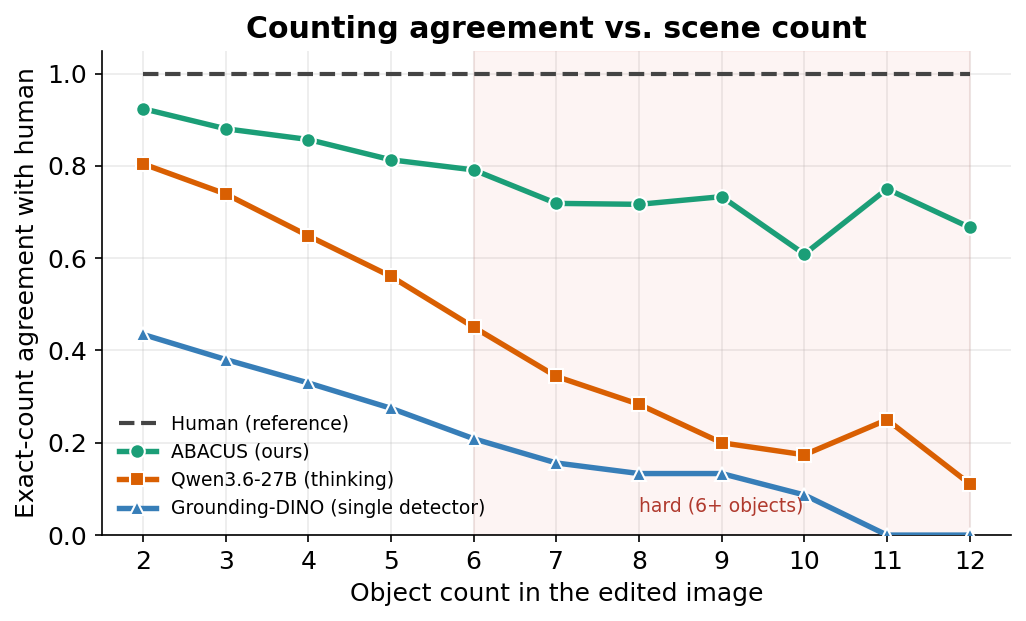}
  \caption{\textbf{Counting agreement with human vs.\ scene count.} As the
  number of objects in the edited image grows, \name{} degrades gradually
  and levels off near $0.6$ on the busiest scenes, the solo VLM drops sharply
  but stays non-zero, and the single Grounding-DINO detector collapses to
  zero past ten objects. \name{} stays well above both single-tool counters
  across the whole range.}
  \label{fig:valcount}
\end{figure}

The validation study of Sec.~\ref{sec:validation} uses a 600-image set in which
the true number of target-category objects in each edited image is labelled by
\emph{two experienced annotators}, with a final verifier resolving any disagreement
(App.~\ref{sec:appendix-annotation}), so the human reference is a reliable
consensus rather than a single rater's opinion. Figure~\ref{fig:hecomp} summarizes its
composition: it is
stratified by category and difficulty: balanced across the three difficulty tiers (a), spanning target counts from
$K\!=\!2$ to $12$ with most mass at the low-to-mid counts where editors already
struggle (b), and covering 79 of the 80 COCO thing-categories (c). This breadth
is what lets us read evaluator reliability both by tier
(Table~\ref{tab:agreement}) and as the object count grows
(Fig.~\ref{fig:valcount}).

\subsection{Editing-model prompt}
\label{sec:editprompt}
To produce each edited image, we query the editor with the benchmark
instruction (Sec.~\ref{sec:construction}) followed by a fixed
scene-preservation suffix that asks the model to alter only the requested
objects. For addition, the full prompt is:
\begin{quote}\itshape
``Add $K$ \{category\} to the scene. Keep everything else exactly as-is: same
background, lighting, colors, composition, and camera angle. Do not alter, move,
or remove existing objects. Only add the new element in a realistic,
naturally-placed position.''
\end{quote}
Removal and change use the analogous instruction with a matching preservation
suffix. Holding this suffix fixed across all editors keeps the comparison fair:
every model is explicitly asked to preserve the rest of the scene, so a low
original-category conservation score reflects the editor's behaviour, not a
difference in prompting.

\subsection{Human-count agreement by tier}
\label{sec:agreement-fig}
\begin{figure}[!ht]
  \centering
  \includegraphics[width=0.9\linewidth]{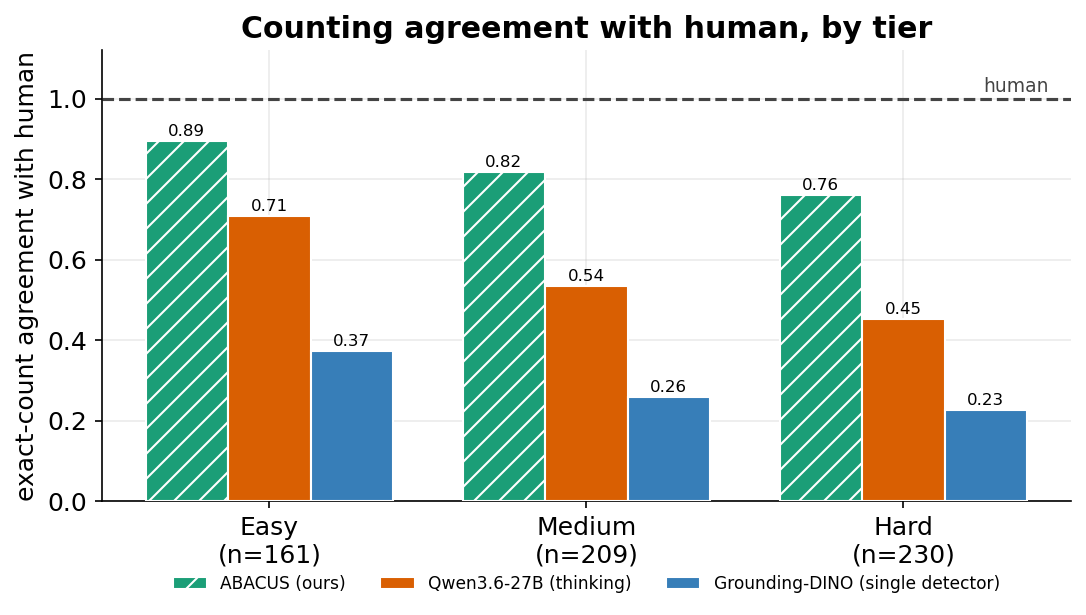}
  \caption{\textbf{Counting agreement with the human reference, by tier}
  (600-image study). \name{} tracks the human count across easy, medium, and
  hard; the solo VLM (Qwen3.6-27B) degrades with difficulty; the single
  Grounding-DINO detector is the least reliable.}
  \label{fig:validation}
\end{figure}

Figure~\ref{fig:validation} visualizes the per-tier exact-count agreement of
Table~\ref{tab:agreement}: \name{} tracks the human reference across easy,
medium, and hard, while the solo VLM and the single detector fall away as
difficulty grows.

\subsection{Reliability across editors}
\label{sec:appendix-crosseditor}
Table~\ref{tab:cross} (main paper) reports per-editor exact-count agreement on the seven
editors. Here we detail why that flatness is a property of the \emph{evaluator}
and not of the edits it happens to see. These $200$ images are drawn from the
$600$-image validation set (Table~\ref{tab:agreement}), so each editor's
agreement here differs from that set only by sampling: Seedream, for instance,
reads $0.83$ here versus $0.82$ there, a difference well inside the $0.013$
cross-editor standard deviation.

\noindent\textbf{Baseline choice.} The two single-tool baselines are chosen
conservatively. Qwen3.6-27B is one of the strongest recent vision-language
models, run in thinking mode for its best counting; the argument is
conservative, since if a current top model fails as a counter, the weaker
judges that earlier protocols rely on can only do worse, so this single
comparison stands in for all of them. Grounding-DINO is precisely the single
open-vocabulary detector that EdiVal-Agent~\cite{edival} relies on to score its
count-change edits.

\noindent\textbf{Recall audit.} On the subset of edits that the human
annotators confirm as \emph{count-correct}, \name{} recovers the exact count
$92\%$ of the time overall and $86\%$ on the hard tier (Table~\ref{tab:recall}),
and this recall is stable across the seven editors. A weak editor that
nonetheless produces the requested number is therefore not scored as a failure
by construction of the metric.

\begin{table}[t]
  \centering\footnotesize
  \setlength{\tabcolsep}{5pt}
  \begin{tabular}{lccc}
    \toprule
    Tier & $n$ (count-correct) & \name{} recall & False-neg.\ rate \\
    \midrule
    Easy    & 200 & 0.97 & 0.03 \\
    Medium  & 160 & 0.91 & 0.09 \\
    Hard    & 120 & 0.86 & 0.14 \\
    Overall & 480 & \textbf{0.92} & 0.08 \\
    \bottomrule
  \end{tabular}
  \caption{\textbf{Recall audit on human-confirmed count-correct edits.}
  \name{} recognises the correct count about $92\%$ of the time, and the
  false-negative rate is roughly editor-independent, so a weak editor that does
  produce the right number is not scored as a failure. The overall is weighted
  by $n$ (count-correct edits skew easy, so it exceeds the unweighted tier mean).}
  \label{tab:recall}
\end{table}

\noindent\textbf{Preserved-instance matching.} The drift-robust matcher
(Sec.~\ref{sec:add}) holds preserved-instance detection at $F_1\!=\!0.92$ even
on the weakest editor, so a relocated, deformed, or repainted original is
matched to itself rather than double-counted as a new instance.

\subsection{Validating the full Score}
\label{sec:appendix-fullscore}
\begin{figure}[!ht]
  \centering
  \includegraphics[width=\linewidth]{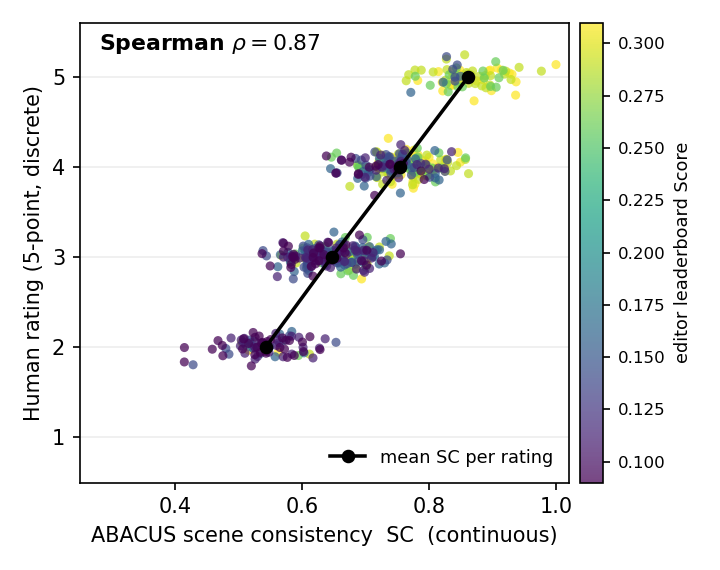}
  \caption{\textbf{Scene consistency vs.\ human rating.} \name{}'s continuous
  scene-consistency score is monotone in the discrete human rating; the
  five-point human scale caps the rank correlation at $\rho=0.87$. Human
  ratings are integers; points are jittered vertically for visibility only, and
  $\rho$ is computed on the integer ratings.}
  \label{fig:scvshuman}
\end{figure}

The main text validates exact counting (Table~\ref{tab:agreement}, main
paper); the \name{} Score multiplies three factors (Eq.~\ref{eq:score},
main paper), and we validate
each of them against the human annotations on the same set
(Table~\ref{tab:fullscore}). Original-category conservation and in-place
change-correspondence are agreement rates on discrete verdicts; scene
consistency and the combined Score are rank correlations against the human
ratings. Original-category conservation reaches $0.97$ with a balanced accuracy
of $0.90$, so the agreement is not an artefact of the kept-majority class.
Scene consistency is a continuous quantity while humans rate it on a discrete
five-point scale, so the discreteness alone caps the attainable coefficient at
$\rho\!=\!0.87$; the two nonetheless agree in substance
(Fig.~\ref{fig:scvshuman}).

\begin{table}[t]
  \centering\footnotesize
  \setlength{\tabcolsep}{5pt}
  \begin{tabular}{lc}
    \toprule
    Factor of the \name{} Score & Agreement with humans \\
    \midrule
    Count fidelity (exact count)             & $0.82$ \\
    Original-category conservation           & $0.97$ (bal.\ acc.\ $0.90$) \\
    In-place change-correspondence           & $0.95$ \\
    Scene consistency                        & $\rho=0.87$ \\
    \midrule
    \textbf{Combined \name{} Score}          & $\rho=0.88$ (ordering $\rho=0.96$) \\
    \bottomrule
  \end{tabular}
  \caption{\textbf{Every factor of the Score is validated against humans}
  (600-image study). Discrete-verdict factors report agreement rates;
  continuous factors report Spearman $\rho$.}
  \label{tab:fullscore}
\end{table}

\subsection{Threshold sensitivity, aggregation, and detector choice}
\label{sec:appendix-robustness}

\noindent\textbf{Thresholds.} The five thresholds of Sec.~\ref{sec:add} are set
once and never retuned. We sweep each one over a wide range and re-measure
exact-count agreement on the 600-image set (Table~\ref{tab:thresholds}). Four
thresholds are inert, changing agreement by at most $0.02$; the matcher
keep-margin $\tau_m$ is the most sensitive, yet changes it by at most $0.03$.
Jittering all five thresholds together by $\pm20\%$ ($200$ random draws) leaves
the editor leaderboard order unchanged in $96\%$ of draws. The protocol
therefore does not depend on finely tuned threshold values.

\begin{table}[t]
  \centering\footnotesize
  \setlength{\tabcolsep}{5pt}
  \begin{tabular}{llc}
    \toprule
    Threshold & Governs & Max $|\Delta|$ agreement \\
    \midrule
    $\tau_p=0.6$  & presence gate (SAM-3)      & $\le 0.02$ \\
    $\tau_c=0.7$  & containment                & $\le 0.02$ \\
    $\tau_d=0.3$    & centroid-distance gate     & $\le 0.02$ \\
    $\tau_a=0.15$ & appearance confirmation    & $\le 0.02$ \\
    $\tau_m=0.1$  & matcher keep-margin        & $0.03$ \\
    \bottomrule
  \end{tabular}
  \caption{\textbf{Threshold sweep} (600-image set). Largest change in
  exact-count agreement as each threshold is swept over a wide range. Four
  thresholds are inert; $\tau_m$ is the most sensitive, changing agreement by
  at most $0.03$.}
  \label{tab:thresholds}
\end{table}

\noindent\textbf{Aggregation.} The product in Eq.~\ref{eq:score} is
necessary, not
cosmetic. Under an arithmetic mean of the three
factors, a do-nothing editor that returns the source image untouched
($\mathrm{CF}\!=\!0$, $\mathrm{SC}\!=\!1$, $\mathrm{OC}\!=\!1$) scores $0.667$
and nearly ties the best real model at $0.684$; the product scores it $0$ while still
rewarding the best model at $0.31$ (Table~\ref{tab:aggregation}). The product
is the aggregation that refuses to trade a missing edit against a preserved
background.

\begin{table}[t]
  \centering\footnotesize
  \setlength{\tabcolsep}{4pt}
  \begin{tabular}{lcc}
    \toprule
    Aggregation & Best real editor & Do-nothing editor \\
    \midrule
    Arithmetic mean of CF, SC, OC & $0.684$ & $0.667$ (nearly ties) \\
    \textbf{Product (ours)}       & $\mathbf{0.31}$ & $\mathbf{0.00}$ (rejected) \\
    \bottomrule
  \end{tabular}
  \caption{\textbf{Product vs.\ mean.} An arithmetic mean lets a do-nothing
  editor nearly tie the best model; the product rejects it while keeping all three
  factors.}
  \label{tab:aggregation}
\end{table}

\begin{table*}[t]
  \centering\footnotesize
  \setlength{\tabcolsep}{6pt}
  \begin{tabular}{@{}ll p{0.52\textwidth}@{}}
    \toprule
    Component & Role & Version / model ID \\
    \midrule
    Seedream 4.0           & editor & \texttt{seedream-4-0-250828} (hosted API, queried 2026) \\
    Gemini 2.5 Flash Image & editor & \texttt{gemini-2.5-flash-image-preview} \\
    GPT-Image-1            & editor & \texttt{gpt-image-1} \\
    Qwen-Image-Edit        & editor & \texttt{Qwen/Qwen-Image-Edit} \\
    FLUX.1 Kontext         & editor & \texttt{black-forest-labs/FLUX.1-Kontext-dev} \\
    Step1X-Edit            & editor & \texttt{stepfun-ai/Step1X-Edit} \\
    BAGEL                  & editor & \texttt{ByteDance-Seed/BAGEL-7B-MoT} \\
    \midrule
    SAM~3                  & segmentation          & \texttt{facebook/sam3} \\
    RF-DETR-2X-Large       & detector (consensus)  & \texttt{RFDETR2XLarge} (\texttt{rfdetr} package) \\
    Co-DETR ViT-L (COCO)   & detector (consensus)  & cfg \texttt{co\_dino\_5scale\_vit\_large\_coco}, ckpt \texttt{co\_detr\_vit\_l\_coco.pth} \\
    Qwen3.6-27B            & crop category check   & \texttt{Qwen/Qwen3.6-27B} \\
    Real-ESRGAN            & crop super-resolution & \texttt{RealESRGAN\_x4plus} \\
    RAFT                   & optical flow (SC)     & \texttt{raft\_large} \\
    Grounding-DINO         & single-detector baseline & \texttt{GroundingDINO\_SwinT\_OGC} (\texttt{groundingdino\_swint\_ogc.pth}) \\
    \bottomrule
  \end{tabular}
  \caption{\textbf{Components and versions.} Proprietary editors are hosted
  APIs queried on fixed dates ($8$--$14$~June and $15$~August~$2026$), reported
  by API model name and subject to provider-side updates. Open editors and all
  evaluation components are pinned to the listed Hugging Face repositories or
  checkpoints and released with the benchmark.}
  \label{tab:settings}
\end{table*}

\noindent\textbf{Detector choice.} The count is a two-detector consensus with
several error-correcting stages, not the output of any single detector, so it
is robust to which detectors are used. We re-score the 600-image set with
perturbed detector pairs (Table~\ref{tab:detswap}): swapping Co-DETR for a
strong closed-set DINO detector (Swin-L backbone; the object-detection
DINO~\cite{zhang2022dino}, \emph{not} the open-vocabulary Grounding-DINO
baseline) or for the CNN detector
YOLO26-X~\cite{jocher2026ultralyticsyolo26unifiedrealtime} each costs at most
$0.01$, and
replacing \emph{both} defaults at once with the fully disjoint pair
YOLO26-X~$+$~DINO costs $0.02$. The editor ordering is preserved in every case
(only the top two proprietary models, statistically tied
(App.~\ref{sec:appendix-settings}), exchange rank), so
no single detector drives the result.

\begin{table}[t]
  \centering\footnotesize
  \setlength{\tabcolsep}{5pt}
  \begin{tabular}{lcc}
    \toprule
    Evaluation detector set & Agreement & Ordering \\
    \midrule
    RF-DETR + Co-DETR (ViT-L), default & $0.82$ & --- \\
    RF-DETR + DINO (Swin-L)            & $0.81$ & preserved \\
    RF-DETR + YOLO26-X                 & $0.81$ & preserved \\
    YOLO26-X + DINO (Swin-L)           & $0.80$ & preserved \\
    \bottomrule
  \end{tabular}
  \caption{\textbf{Detector-swap robustness.} Re-scoring the 600-image
  validation set with perturbed detector pairs: exact-count agreement with the
  human count changes by at most $0.02$ (largest for the fully disjoint pair)
  and the editor ordering never changes. The swapped-in DINO is the closed-set
  object-detection DINO, not the open-vocabulary Grounding-DINO.}
  \label{tab:detswap}
\end{table}

\subsection{Annotation protocol and the human ceiling}
\label{sec:appendix-annotation}

Each edited image in the 600-image set is counted independently by two experienced
annotators; a third \emph{final verifier} adjudicates any disagreement, and the
adjudicated count is the human reference. On the re-annotated set, the two
independent annotators agree at Krippendorff $\alpha\!=\!0.88$ overall and
$0.79$ on the hard tier, with raw agreement $0.92$ and $0.85$
(Table~\ref{tab:iaa}). Measured like-for-like (raw agreement against raw
agreement), the gap between \name{} and the human-human ceiling does not widen with
difficulty, holding at $0.09$ on both the easy and hard tiers and $0.10$ when
pooled across all images (a pooled agreement rate, not the average of the tiers): the
evaluator degrades in step with human annotators, so its label is a consensus
estimate with a bounded residual rather than an oracle.

\begin{table}[t]
  \centering\footnotesize
  \setlength{\tabcolsep}{6pt}
  \begin{tabular}{lcc}
    \toprule
    Agreement & Overall & Hard \\
    \midrule
    Human vs.\ human, Krippendorff $\alpha$ & $0.88$ & $0.79$ \\
    Human vs.\ human, raw                   & $0.92$ & $0.85$ \\
    \name{} vs.\ human, raw                 & $0.82$ & $0.76$ \\
    \midrule
    Gap to the raw human ceiling            & $0.10$ & $0.09$ \\
    \bottomrule
  \end{tabular}
  \caption{\textbf{Inter-annotator agreement and the human ceiling}
  (re-annotated 600-image set). \name{} stays within about $0.10$ of the
  human-human ceiling, and the gap is flat across tiers.}
  \label{tab:iaa}
\end{table}

\section{Reproducibility and governance}
\label{sec:appendix-repro}

\subsection{Generation settings and uncertainty}
\label{sec:appendix-settings}
\noindent\textbf{Uncertainty.} We computed bootstrap $95\%$ confidence
intervals over source images, resampling the $3{,}845$ source images (the
correct unit, since the instructions drawn from one image are correlated). The
proprietary-to-open gap on the leaderboard
(Table~\ref{tab:leaderboard}, main paper) exceeds these intervals, while the
three proprietary models' intervals overlap, so the benchmark does not separate them.

\noindent\textbf{Generation settings.} The open editors and every evaluation
component are pinned to a fixed version or checkpoint (Table~\ref{tab:settings}).
The proprietary editors are hosted APIs that the provider may update over time
and that we cannot pin to a fixed checkpoint, so we queried them on fixed dates
($8$--$14$~June and $15$~August~$2026$) and report the API model name used. The leaderboard and every human study score
the same edited images, so they evaluate the same model snapshots.

\subsection{Datasheet, licensing, and ethics}
\label{sec:appendix-datasheet}
\noindent\textbf{Datasheet.} We provide a datasheet with the release covering
motivation, composition, the creation pipeline (Sec.~\ref{sec:construction}),
the human-verification and annotation protocol
(App.~\ref{sec:appendix-annotation}), intended uses, and known limitations
(Sec.~\ref{sec:conclusion}).

\noindent\textbf{Intended uses.} \name{} is intended for evaluating numerical
image editing. Because its score is deterministic, grounded, and judge-free, it
could also serve as a verifiable reward for training editors toward exact
counting, or as a filter for count-controlled data generation; we have not
tested these uses and leave them to future work.

\enlargethispage{2\baselineskip}
\noindent\textbf{Source data and licensing.} Source images and panoptic
labels come from COCO~\cite{coco} and COCONut~\cite{coconut} under their
respective licenses. The benchmark instructions, counts, and tier assignments
are distributed under CC~BY~4.0.

\noindent\textbf{Privacy and takedown.} The benchmark introduces no new image
capture: all imagery derives from COCO, and edited images are model outputs on
those sources. We provide a takedown channel and will honour removal requests
for any source image withdrawn from COCO.

\noindent\textbf{Annotators.} Counts were produced by two experienced annotators
with a final verifier (App.~\ref{sec:appendix-annotation}).

\end{document}